\documentclass[journal]{IEEEtran}

\usepackage{subfigure}
\usepackage{multirow,}
\usepackage{booktabs}
\usepackage{bm}
\usepackage{xcolor}
\usepackage{color, soul}
\definecolor{codegray}{rgb}{0.5,0.5,0.5}

\usepackage{enumitem}
\usepackage{graphicx}
\usepackage{algorithmic}
\usepackage[ruled,vlined,linesnumbered]{algorithm2e}
\usepackage{enumitem}
\usepackage{bbm}
\usepackage{amsfonts}
\usepackage{amsmath}
\usepackage{cite}
\usepackage{subcaption}  

\usepackage{hyperref}
\hypersetup{
    colorlinks=true,
    linkcolor=green,
    citecolor=green,
    urlcolor=blue
}
\usepackage{color, soul}

\begin{document}
\title{RobGC: Towards Robust Graph Condensation}

\author{Xinyi~Gao,~\IEEEmembership{Student Member,~IEEE,}
Hongzhi Yin,~\IEEEmembership{Senior Member,~IEEE,}
Tong Chen,~\IEEEmembership{Member,~IEEE,} \\
Guanhua Ye,
Wentao Zhang,
Bin Cui,~\IEEEmembership{Fellow,~IEEE}

\thanks{The Australian Research Council partially supports this work under the streams of Future Fellowship (Grant No. FT210100624), Discovery Early Career Researcher Award (Grants No. DE230101033), the Discovery Project (Grant No. DP240101108 and DP240101814), and the Linkage Projects (Grant No. LP230200892 and LP240200546).}

\thanks{Xinyi Gao, Hongzhi Yin, and Tong Chen are affiliated with the School of Electrical Engineering and Computer Science, the University of Queensland, Brisbane, Australia.
Guanhua Ye is with the School of Computer Science, Beijing University of Posts and Telecommunications, Beijing, China. 
Wentao Zhang is with the Center for Machine Learning Research, Peking University, Beijing, China. 
Bin Cui is with the School of Computer Science, Peking University, Beijing, China.}
\thanks{Corresponding author: Hongzhi Yin (e-mail: h.yin1@uq.edu.au).}

}

\markboth{Journal of \LaTeX\ Class Files,~Vol.~14, No.~8, August~2015}%
{Xinyi Gao \MakeLowercase{\textit{et al.}}: Bare Demo of IEEEtran.cls for IEEE Journals}

\maketitle
\begin{abstract}
The increasing prevalence of large-scale graphs presents a significant challenge for graph neural networks (GNNs) training due to their computational demands, limiting the applicability of GNNs in various scenarios. In response to this challenge, graph condensation (GC) is proposed as a promising acceleration solution, focusing on generating an informative compact graph that enables efficient training of GNNs while retaining performance. Despite the potential to accelerate GNN training, existing GC methods overlook the quality of large training graphs during both the training and inference stages. They indiscriminately emulate the training graph distributions, making the condensed graphs susceptible to noises within the training graph and significantly impeding the application of GC in intricate real-world scenarios. To address this issue, we propose robust graph condensation (RobGC), a plug-and-play approach for GC to extend the robustness and applicability of condensed graphs in noisy graph structure environments. Specifically, RobGC leverages the condensed graph as a feedback signal to guide the denoising process on the original training graph. A label propagation-based alternating optimization strategy is in place for the condensation and denoising processes, contributing to the mutual purification of the condensed graph and training graph. Additionally, as a GC method designed for inductive graph inference, RobGC facilitates test-time graph denoising by leveraging the noise-free condensed graph to calibrate the structure of the test graph. Extensive experiments show that RobGC is compatible with various GC methods, significantly boosting their robustness. 
\end{abstract}


\IEEEpeerreviewmaketitle

\section{Introduction}

\begin{figure}[t]
\setlength{\abovecaptionskip}{0.1cm}
\centering
\includegraphics[width=\linewidth]{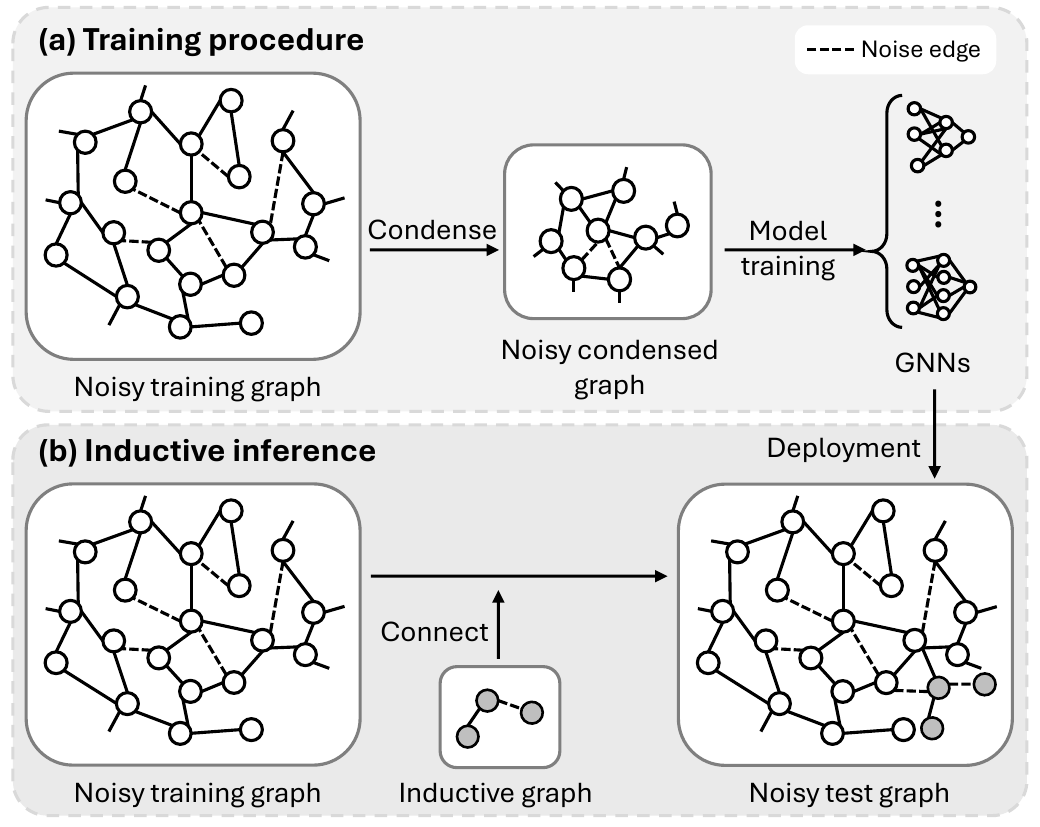}
\caption{The graph condensation process and inductive inference scenario within the noisy environment. (a) In the training procedure, the noisy training graph is condensed and then the condensed graph is leveraged to train multiple GNN models with different architectures. (b) In the inference stage, new graph is integrated into the noisy training graph, and GNNs are then deployed on this expanded noisy graph for inference.}
\label{fig_pre}
\end{figure}

\IEEEPARstart{G}{raph} data \cite{zheng2016keyword,gao2023semantic,gao2023accelerating} is extensively utilized across diverse domains, owing to its capability to represent complex structural relationships among various entities in the real world. To address the imperative for data management, graph neural networks (GNNs) have attracted significant attention for their exceptional representation capabilities. By incorporating node attributes and graph topology, they could extract latent information within graphs and have been utilized in a wide range of real-world applications, including chemical molecules \cite{guo2022graph}, social networks \cite{li2018influence}, and recommender systems \cite{jiang2024challenging}.
However, the burgeoning volume of graph data within these real-world applications poses formidable challenges in training GNN models.
GNNs commonly employ the message-passing mechanism that propagates node features across the graph structure. Despite its effectiveness, this method can lead to the neighbor explosion issue \cite{DBLP:conf/iclr/ZengZSKP20}, which considerably escalates the computational requirements during the model training stage. 
This situation becomes even more challenging when training multiple GNN models,
such as in the application of neural architecture search~\cite{zhang2022pasca}, continual learning~\cite{rebuffi2017icarl}, and federated learning~\cite{pan_fedgkd_2023}.

In response to the urgent demand for processing large-scale graph data, graph condensation (GC) \cite{jin2022graph,gao2024graph} is proposed to substitute the large training graph with a compact (e.g., $1,000\times$ smaller) yet informative graph to accelerate the GNN training.
By retaining the crucial characteristics of the large training graph, GNN models trained on the condensed graph demonstrate comparable performance to those trained on the large training graph, while significantly reducing model training time.
This not only facilitates large-scale graph management but also expands the application of GNNs in resource-constrained scenarios for various downstream tasks.
{{Among these, we specifically focus in this paper on the \textit{inductive inference task}, where new and previously unseen nodes are continuously integrated into the existing training graph as depicted in Fig. \ref{fig_pre}. 
This task is practical and widely adopted in high-throughput graph systems~\cite{van2022inductive, hung2017computing}. For instance, online social networks~\cite{nguyen2017argument} frequently integrate new posts and users into their existing graphs for trend monitoring and anomaly detection. Large-scale e-commerce platforms~\cite{wang2018streaming} incorporate new products and users into their recommender systems to improve cold-start recommendations and adapt to evolving product landscapes. By integrating with GC, GNNs trained on condensed graphs can be deployed on these expanded graphs to generate embeddings for unseen nodes and make predictions~\cite{hamilton2017inductive}.}}

\begin{figure}[t]
\setlength{\abovecaptionskip}{0.1cm}
\centering
\begin{minipage}[t]{0.465\linewidth}
\centering
\includegraphics[width=\linewidth]{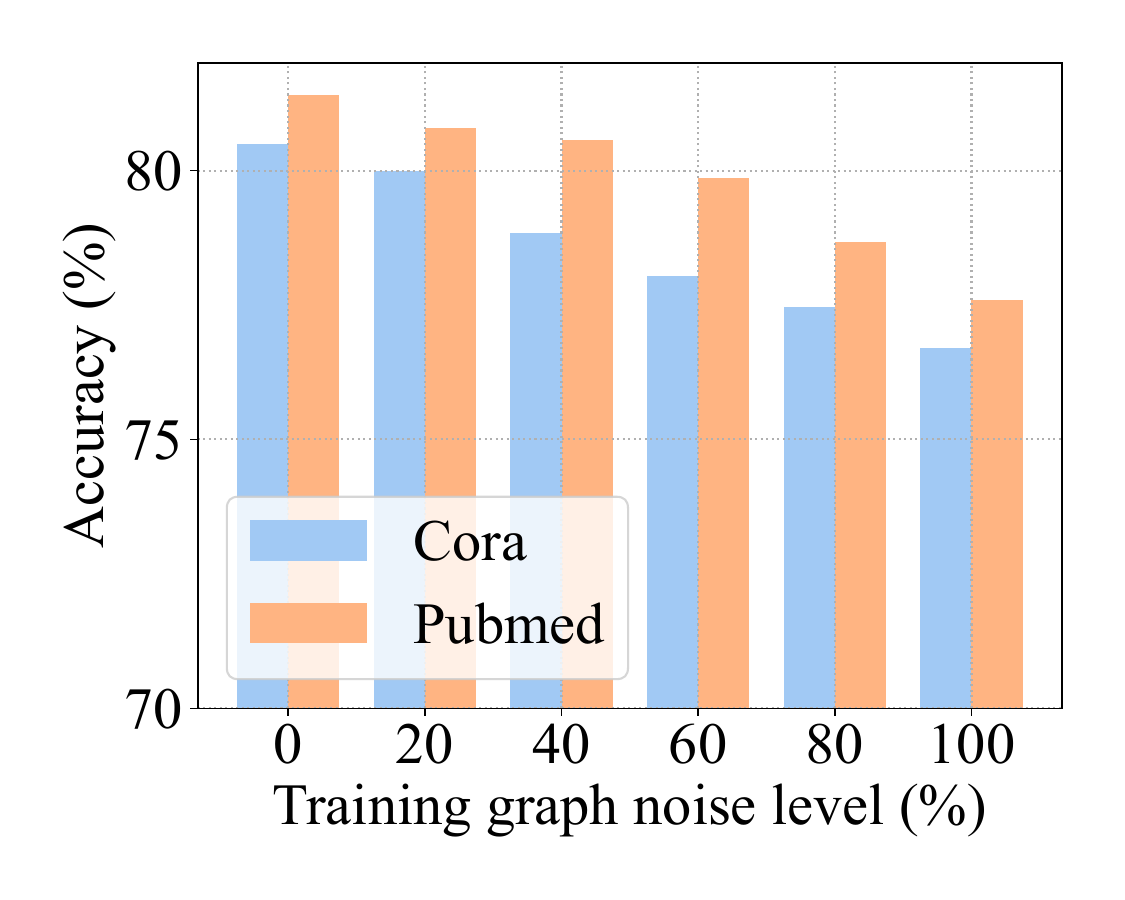}
\end{minipage}
\hfill
\begin{minipage}[t]{0.465\linewidth}
\centering
\includegraphics[width=\linewidth]{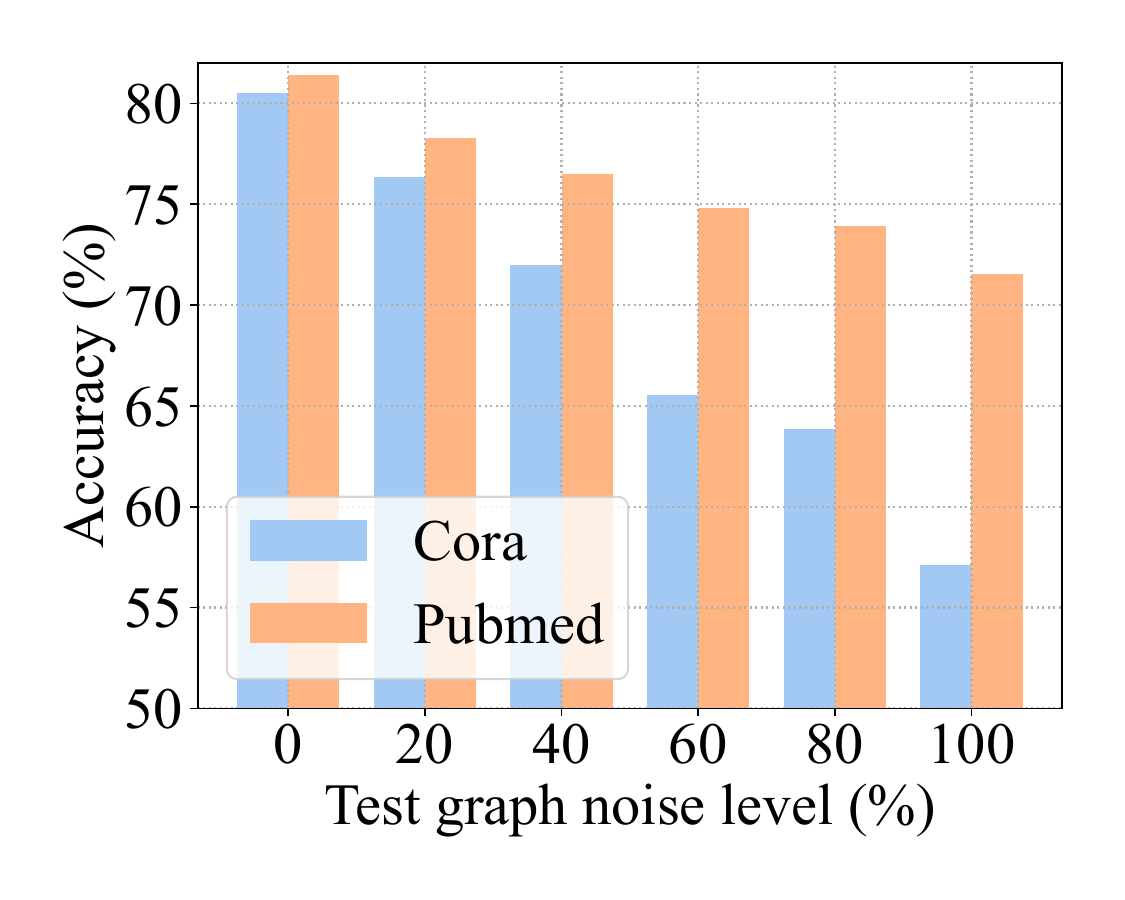}
\end{minipage}
\caption{The test accuracy of GCN~\cite{DBLP:conf/iclr/KipfW17} trained on the condensed graphs. GCond~\cite{jin2022graph} is adopted to condense the graph and random noise is applied.
(a) The performance of the GCN trained on condensed graphs derived from the training graph with different levels of noise and tested on a clean graph. (b) The results of the GCN trained on a clean condensed graph and tested on graphs containing varying levels of noise.}
\label{fig_obs}
\end{figure}

Despite the potential of GC methods to expedite model training, their practical applicability remains sub-optimal in intricate real-world scenarios.
Predominantly, existing GC methods presuppose the presence of clean training graph structures, contradicting the complicated nature of data processing in practical scenarios.
Unfortunately, real-world graph structures inevitably contain noisy edges due to data collection errors \cite{randall2014use}, outdated information \cite{tu2023deep} or inherent variability in dynamic environments \cite{gao2021training}. These unexpected noises significantly impacts the effectiveness of GC in both the condensation and inference stages.
As depicted in Fig. \ref{fig_pre} (a), given the large training graph, the objective of the GC in condensation stage is indiscriminately emulating the distribution of the training graph, disregarding the intrinsic quality of the training graph. 
As a result, the condensed graph inherits the noisy distribution from the training graph, which is carried over to GNNs and compromises the model prediction accuracy \cite{geisler2021robustness}.
In addition, the representation of inductive nodes in the inference stage is more challenging with structural noises.
As shown in Fig. \ref{fig_pre} (b), inductive nodes for prediction are integrated into the large training graph only during the test time. 
Consequently, achieving high-quality representations of these inductive nodes becomes imperative for ensuring accurate predictions. 
However, the presence of noisy structures within both the training and inductive graphs propagates noisy information to these inductive nodes, exacerbating the challenge of precise representation for them. 
To assess the effect of noisy structures on GC, we simulate the deployment of GC under noisy training and test graphs separately.
As depicted in Fig. \ref{fig_obs} (a), the performance decline on the clean test data indicates that the noisy structures in the training graph severely disrupt the graph distributions and are indiscriminately simulated by condensed graphs.
Moreover, when GNNs are trained on a clean condensed graph but evaluated on the noisy inductive test graphs, the performance decline is more pronounced as shown in Fig. \ref{fig_obs} (b), underscoring the detrimental effect of noisy structures on inductive node representation.
In a nutshell, the critical problem that arises for GC is: ``\textit{How can GC methods effectively distill essential knowledge from noisy training graph into a condensed graph, while ensuring that GNNs trained on these condensed graphs maintain robust against noisy inductive test graphs?}''

To empower GC with robustness against both noisy training and test graph structure, we aim to introduce a general graph structure optimization procedure in both the condensation and inference stages.
Specifically, the desiderata of robust GC under the inductive setting is: 
(1) during condensation, the condensed graph extracted from the noisy training graph is noise-free, thus containing only the core, causal information for effective GNN training; 
(2) during inference, an efficient mechanism is in place to denoise the inductive graph that is only seen during test time, so as to prepare a clean graph for GNN inference on-the-fly.
To do so, the first challenge comes from jointly optimizing the denoising procedure during the condensation stage.
A potential solution for this challenge is graph structure learning (GSL) \cite{zhu2021deep}, employing a specific GSL model to optimize the graph structural and learn denoised node representations.
However, these GSL methods tightly entangle model training with graph structure optimization, thereby inevitably complicating GC optimization by introducing the additional model training task. 
Therefore, it is necessary to design an efficient and applicable optimization strategy for the denoising process.
In the meantime, the second challenge involves the efficient inductive graph denoising during the inference stage. 
Existing GSL approaches optimize the graph structure alongside the GNN model, which restricts their capability to handle nodes observed solely during the training stage.
Consequently, when inductive nodes are introduced to the training graph, these methods require retraining the model and optimize the graph from scratch, contradicting the purpose of employing GC for efficiency.
Ideally, the solution would involve learning a principled denoising function during the condensation stage, capable of denoising the inductive test graph without the need for retraining. 
{{Besides GSL, another related area is robust GNNs~\cite{geisler2020reliable}, which focus on developing specialized modules to enhance the robustness of models against noise~\cite{li2022reliable} and adversarial attacks~\cite{geisler2021robustness}. However, these model-centric approaches are constrained by the specific model architecture and downstream task, and fall short of addressing the deficiencies inherent in the data itself.}}

In light of these challenges, we introduce the \underline{Rob}ust \underline{g}raph \underline{c}ondensation (RobGC), a model-agnostic approach for GC to extend the robustness and applicability of condensed graphs in noisy scenarios.
Instead of decoupling the design of the condensation process and inductive graph denoising process, we innovatively bridge the first and second desiderata using the condensed graph. 
In our proposed RobGC, when learning the condensed graph, we additionally require it to serve the purpose of amending the noisy structure of the training graph. 
This ensures that the mutual promotion of the quality of the training and condensed graph, and the condensed graph only contains the underlying information of the training graph. 
As such, when given an inductive graph that is also subject to structural noise, we can use the noise-free condensed graph as a reference point to denoise it with a considerably low computational cost.
Specifically, we formulate the correlation between the training and condensed graphs to assess the reliability of edges and adopt label propagation as the optimization objective. 
The refinement of the graph structure is regulated by thresholds, which can be efficiently determined through grid search and deployed in the inference stage for test-time denoising. 
This training-free paradigm eliminates any GNN models in structure optimization and ensures architectural independence and scalability of the denoising procedure. 
Consequently, RobGC is well-suited for managing both noisy training and test graph structures to generate high-quality condensed graphs and improve prediction performance.
The main contributions of this paper are threefold:

\begin{itemize}[leftmargin=*]
\item  We are the first (to the best of our knowledge) to focus on the data quality issue of GC in the noisy scenario and point out the necessity of identifying the noise structure in the training and test graph, which is an important yet under-explored problem in GC.
\item  We present RobGC, a plug-and-play approach for GC methods that enhances the robustness of condensed graphs within noisy environments. 
RobGC adopts the condensed graph as the denoising signal for both training and test graph and leverages label propagation for efficient structure optimization. 
The processes of condensation and structure optimization are conducted interactively, contributing to the mutual promotion of the quality of the training graph and condensed graph.
\item 
Through extensive experimentation, we verify that RobGC is compatible with different GC methods. It excels in handling various types and levels of graph structure noise, all while preserving the inherent generalization capabilities of GC. { The code is available at: \href{https://github.com/XYGaoG/RobGC}{https://github.com/XYGaoG/RobGC}. }
\end{itemize}

\section{Preliminaries}
In this section, we first introduce the GNN and conventional GC methods, and then formally define the problem studied.

\subsection{Graph Neural Networks}
\label{sec_gcn}
Consider that we have a large-scale graph $\mathcal{T}=\{{\bf A}, {\bf X}\}$ consisting of $N$ nodes. 
${\bf X}\in{\mathbb{R}^{N\times d}}$ denotes the $d$-dimensional node feature matrix and ${\bf A}\in \mathbb{R}^{N\times N}$ is the adjacency matrix. The node and edge set are denoted by $\mathcal{V}$ and $\mathcal{E}$, respectively. We use ${\bf Y}\in{\mathbb{R}^{N\times C}}$ to denote the one-hot node labels over $C$ classes.
GNNs learn the embedding for each node by leveraging the graph structure and node features as the input.
Without loss of generality, we use graph convolutional network (GCN) \cite{DBLP:conf/iclr/KipfW17} as an example, where the convolution operation in the $k$-th layer is defined as follows: 
\begin{equation}
{\bf{H}}^{(k)} =\text{ReLU}\left(\hat{\mathbf{A}}{\bf H}^{(k-1)}{\mathbf W}^{(k)}\right),   
\label{eq_gcn}
\end{equation}
where ${\bf{H}}^{(k)}$ is the node embeddings of the $k$-th layer, and ${\bf{H}}^{(0)}=\mathbf{X}$.
$\hat{\mathbf{A}}=\widetilde{\mathbf{D}}^{-\frac{1}{2}}\widetilde{\mathbf{A}}\widetilde{\mathbf{D}}^{-\frac{1}{2}}$ is the normalized adjacency matrix. 
$\widetilde{\mathbf{A}}$ represents the adjacency matrix with the self-loop, $\widetilde{\mathbf{D}}$ denotes the degree matrix of $\widetilde{\mathbf{A}}$, and $\mathbf{W}^{(k)}$ is the trainable weights. 
$\text{ReLU}\left(\cdot\right)$ is the activation function.
In this paper, we concentrate on the node classification task, where the $K$-th layer embeddings ${\bf{H}}^{(K)}$ are further input into the classifier for prediction.

\subsection{Graph Condensation}
\label{sec_gc}
Graph condensation \cite{jin2022graph} aims to generate a small condensed graph $\mathcal{S}=\{{\bf A'}, {\bf X'}\}$ with ${\bf A'}\in\mathbb{R}^{N'\times N'}$, ${\bf X'}\in\mathbb{R}^{N'\times d}$ as well as its label ${\bf Y'}\in{\mathbb{R}^{N'\times C}}$, where $N'\ll{N}$.
GNNs trained on $\mathcal{S}$ can achieve comparable performance to those trained on the much larger $\mathcal{T}$. 
To connect the training graph $\mathcal{T}$ and condensed graph $\mathcal{S}$, a relay model $f_{\theta}$ parameterized by ${\theta}$ is employed in GC for encoding both graphs.
The losses of $\mathcal{T}$ and $\mathcal{S}$ w.r.t. ${\theta}$ are defined as:
\begin{equation}
\label{eq_iniloss}
\begin{split}
&\mathcal{L} \left( \theta \right) = \ell \left( f_{\theta} \left( {\mathcal{T}} \right) , \mathbf{Y} \right),\\
&\mathcal{L}' \left( \theta \right)  = \ell \left( f_{\theta} \left( {\mathcal{S}} \right) , \mathbf{Y}' \right),  
\end{split}
\end{equation}
where $\ell\left( \cdot,\cdot \right)$ is the classification loss function such as cross-entropy and $\mathbf{Y}'$ is predefined to match the class distribution in $\mathbf{Y}$.
Then the objective of GC can be formulated as a bi-level optimization problem:
\begin{equation}
\label{eq_oriloss}
\min_{\mathcal{S}} \mathcal{L}  \left( \theta^{\mathcal{S}} \right)
\:\: \text{s.t.}  \:\: \theta^{\mathcal{S}}   = \arg\min_{\mathcal{\theta}} \mathcal{L}'  \left( \theta \right) .
\end{equation}
To solve the objective outlined above, GCond~\cite{jin2022graph} proposes to match the model gradients at each training step $t$. In this way, the training trajectory on condensed graph can emulate that on the training graph, i.e., the models trained on these two graphs converge to similar solutions. The objective is formulated as:
\begin{equation}
\label{eq_gmloss}
\begin{split}
& \mathcal{L}_{cond} = \min_{\mathcal{S}} \mathbb{E}_{\theta_0 \sim {\Theta}} \left[ \sum_{t=1}^{T} \mathcal{D} \left( \nabla_{\theta_t}\mathcal{L} \left( {\theta_t}\right), \nabla_{\theta_t}\mathcal{L}' \left( {\theta_t} \right) \right) \right] \,\, \\
& \text{s.t.} \,\, {\theta}_{t+1} = \operatorname{opt}\left( \mathcal{L}'\left( {\theta_t} \right) \right),
\end{split}
\end{equation}
where $\theta_0$ represents the initial parameters of the relay model, which is sampled from the distribution ${\Theta}$. The expectation on $\theta_0$ aims to improve the robustness of ${\mathcal{S}}$ to different parameter initialization \cite{lei_comprehensive_2024}. 
$\operatorname{opt}\left(\cdot\right)$ is the model parameter optimizer and the parameters of the relay model are updated only on $\mathcal{S}$.
$\mathcal{D} \left( \cdot,\cdot \right) $ is the distance measurement to calculate the gradient distances. Suppose the gradient $\nabla_{\theta}\mathcal{L}=\{\mathbf{G}^{(p)}\}_{p=1}^{P}$ and $\nabla_{\theta}\mathcal{L}'=\{\mathbf{G}'^{(p)}\}_{p=1}^{P}$ in Eq. \eqref{eq_gmloss} entails all $P$ layers' model gradient matrices $\mathbf{G}^{(p)}$ and $\mathbf{G}'^{(p)}$, $\mathcal{D} \left( \cdot,\cdot \right) $ is calculated by summing up all layers' pairwise gradient distances: 
\begin{equation}
\mathcal{D}
(\nabla_{\theta}\mathcal{L},\nabla_{\theta}\mathcal{L}')=\sum_{p=1}^{P}\sum_{i=1}^{D_{p}}\left(1-\text{cos}\left(\mathbf{G}^{(p)}_i, {\mathbf{G}'_i}^{(p)}\right)\right),
\end{equation}
where $\text{cos}(\cdot,\cdot)$ is the cosine similarity, $\mathbf{G}^{(p)}_i$ and $\mathbf{G}'^{(p)}_i$ are the $i$-th column vector (out of all $D_{p}$) in the gradient matrix $\mathbf{G}^{(p)}$ and $\mathbf{G}'^{(p)}$ at layer $p$, respectively. 

To simplify the optimization of $\mathcal{S}$, the structure of condensed graph $\mathbf{A}'$ is parameterized by $\mathbf{X}'$ as:
\begin{equation}
\label{eq:adj}
{\bf A}_{i,j}' = \sigma\left(\frac{{\text{MLP}([{\bf X}'_i; {\bf X}'_j])} + {\text{MLP}([{\bf X}'_j; {\bf X}'_i])}}{2}\right),
\end{equation}
where $\text{MLP}(\cdot)$ is a multi-layer perceptron, and fed with the concatenation of condensed node features ${\bf X}'_i$ and ${\bf X}'_j$. $\sigma$ is the sigmoid function.

In addition to gradient matching, a variety of GC methods adopt distinct optimization strategies to address the optimization problem in Eq. (\ref{eq_oriloss}). {{For instance, GCDM \cite{zhao2023dataset} and CAT\cite{liu_cat_2023} introduces the distribution matching, aiming to align the feature distributions between the condensed and training data. GDEM \cite{liu_graph_2023} directly generates the condensed graph's eigenbasis to match that of the training graph. GC-SNTK \cite{wang2023fast} and OpenGC \cite{gao2024grapho} leverages the kernel ridge regression to address the bi-level optimization by the close-form solution.}} Meanwhile, SFGC \cite{zheng_structure_free_2023} and GEOM \cite{zhang2024navigating} brings trajectory matching into the GC field, synchronizing the long-term learning behaviors of GNNs across the training and condensed graphs. Despite their potential to enhance GNN performance, these GC methods neglect the critical aspect of graph quality, which is essential for real-world applications.

\begin{figure*}[ht]
\setlength{\abovecaptionskip}{0.1cm}
\centering
\includegraphics[width=0.9\linewidth]{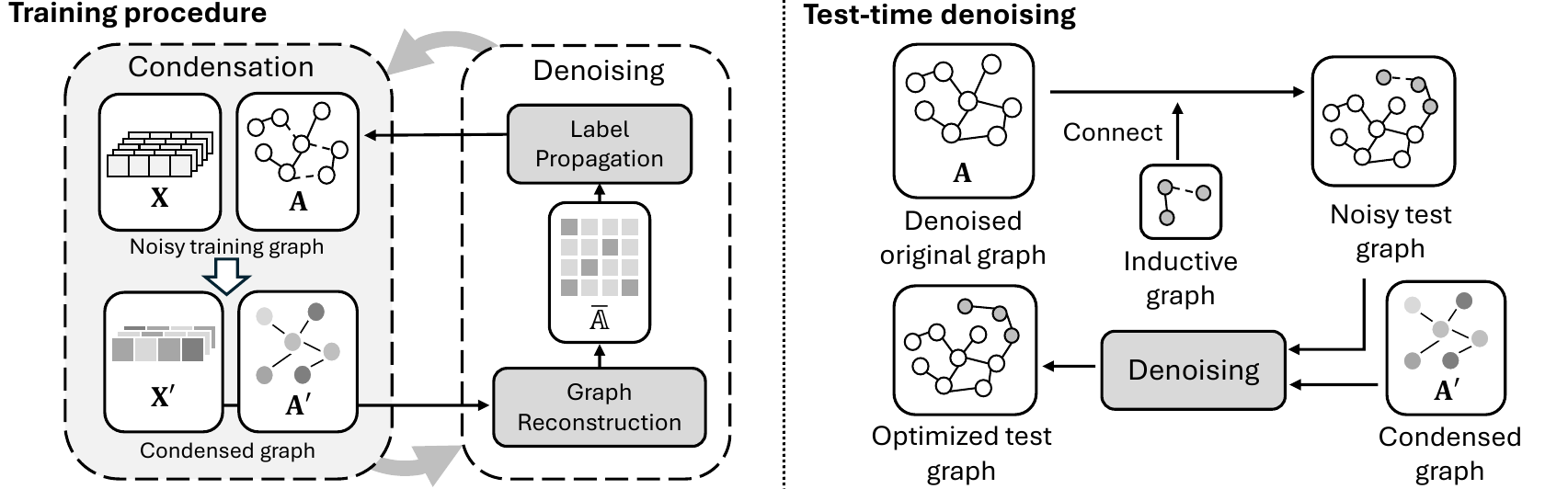}
\caption{{{The training and test procedures of RobGC. During the training stage, the noisy training graph is condensed, and the condensed graph is utilized as a denoising signal to optimize the structure of the training graph. In the inference stage, the noisy test graph is denoised by referring the condensed graph. Features are omitted in test-time denoising for simplicity.}}}
\label{fig_main}
\end{figure*}

\subsection{Problem Formulation}
\label{sec_formulation}

We first formulate the inductive inference task of GC with noisy graph structures, and then define the robust GC problem.

We consider a large graph with noisy and missing edges, partitioned into three non-overlapping subsets: the training graph $\mathcal{T} = \{\mathbf{A}, \mathbf{X}\}$, the validation sub-graph $\mathcal{T}_v$, and the test sub-graph $\mathcal{T}_t$. 
In the GC process, only $\mathcal{T}$ is accessible to generate the condensed graph $\mathcal{S} = \{\mathbf{A}', \mathbf{X}'\}$, which is used to train GNN models. During validation/testing, $\mathcal{T}_v / \mathcal{T}_t$ are integrated into $\mathcal{T}$ to create the validation/test graphs $\mathcal{T}_{val} / \mathcal{T}_{test}$ for performance evaluation.

The goal of robust GC is to generate a clean $\mathcal{S}$ alongside a graph structure optimization function $g: (\mathcal{T}, \mathcal{S}) \to \mathbb{A}$, where $\mathbb{A} \in \mathbb{R}^{N \times N}$ represents the optimized, clean training graph structure of $\mathcal{T}$. Consequently, during inference, $\mathcal{T}_{test}$ can be denoised by $g: (\mathcal{T}_{test}, \mathcal{S}) \to \mathbb{A}_{test}$, and GNNs trained on $\mathcal{S}$ are deployed on $\mathbb{A}_{test}$ for predictions.

\section{Methodologies}
We hereby present our proposed RobGC, which includes the alternating optimization between graph condensation and graph denoising, as well as a test-time denoising strategy, as depicted in Fig. \ref{fig_main}.   
We begin with the training graph denoising by introducing the reliable graph structure modeling approach.
Subsequently, we define the optimization objective and elaborate on the alternating optimization framework in detail. 
Finally, the test-time denoising strategy utilizes the noise-free condensed graph to calibrate the inductive test graph for performance enhancement.

\subsection{Reliable Graph Structure Modeling}

To effectively integrate the condensation and denoising processes at both training and test phases, we utilize the condensed graph as a critical denoising signal. This approach not only bridges these two processes but also provides a holistic overview of the training graph's structure and node features for the denoising process, facilitating a more comprehensive and precise graph structure formulation. 

Specifically, we first compute the similarity between training graph node ${\mathbf{X}_i}$ and condensed graph node ${\mathbf{X}'_j}$ as:
\begin{equation}
\mathbf{E}_{i,j} = \text{cos}({\mathbf{X}_i}, {\mathbf{X}'_j}),
\label{eq_cosine1}
\end{equation}
where $\text{cos}(\cdot,\cdot)$ measures the cosine similarity. $\mathbf{E} \in [-1,1]^{N\times N'}$ explicitly encodes the correlation between the training graph and condensed graph. Specifically, each row $\mathbf{E}_i$ is a $N'$-dimensional vector, where the entries indicate weighted correlations between the training node $i$ and all condensed nodes. 
In other words, each training node $i$ can be represented by the weighted ensemble of condensed nodes identified by $\mathbf{E}_i$. Similarly, due to the symmetric characteristic of cosine similarity, we define the correlations between the condensed nodes and training nodes as $\mathbf{U}=\mathbf{E}^{\mathsf{T}}$, and each condensed graph node $j$ can be represented by training nodes according to $\mathbf{U}_j$.

Then, we enhance the correlation to incorporate high-order relationships by propagating $\mathbf{U}$ on the condensed graph ${\bf A'}$. The propagation is formulated as:
\begin{equation}
{\bf{U}}^{(k)} =\hat{\mathbf{A}}'{\bf U}^{(k-1)},   
\label{eq_propagation}
\end{equation}
where $\hat{\mathbf{A}}'$ is the normalized adjacency matrix of ${\bf A'}$, and ${\bf{U}}^{(0)} = {\bf{U}}$.
This process enables the aggregation of $k$-th order neighbor information captured by the condensed graph, enriching the node representations with broader contextual insights.
With the correlation between training nodes and condensed nodes, we integrate them with node features to assess the reliability of edges between node $i$ and $j$ as:
\begin{equation}
\begin{aligned}
&\bar{\mathbb{A}}_{i,j} = \text{cos}({\mathcal{M}_i}, {\mathcal{W}_j}),\\
&{\mathcal{M}_i} = \left[{\bf X}\mathbin\Vert\mathbf{E} \mathbin\Vert ... \mathbin\Vert \mathbf{E}\right]_i,\\
&{\mathcal{W}_j} = \left[{\bf X} \mathbin\Vert \mathbf{U}^{(0){\mathsf{T}}} \mathbin\Vert ... \mathbin\Vert \mathbf{U}^{(K){\mathsf{T}}}\right]_j,
\end{aligned}
\label{eq_cosine2}
\end{equation}
where $\mathbin\Vert$ denotes the concatenation operation, and the correlation matrix is up to $K$-th order.
Higher node similarity suggests increased reliability of edges and a stronger probability of connection. This conforms to the widely-accepted network homophily assumption \cite{zhu2021deep}, a core principle in graph structure learning \cite{liu2022towards} and GNN design \cite{DBLP:conf/iclr/KipfW17}.

While Eq. (\ref{eq_cosine2}) enables the calculation of a new training adjacency matrix, determining the reliability $\bar{\mathbb{A}}_{i,j}$ for each node pair involves a high computational complexity of $\mathcal{O}(N^2)$, resulting in extensive processing time. To accelerate the optimization procedure, we employ a dual-phase structure modification approach to delete unreliable edges and add the potentially effective edges separately.
Firstly, unreliable edge deletion is performed to assess existing edges within the training graph edge set $\mathcal{E}$, and remove the edges if there reliability is smaller than the threshold $\epsilon_1$:
\begin{equation}
{\mathbb{A}}_{i,j}=\left\{\begin{matrix}
 0 & \bar{\mathbb{A}}_{i,j}\le \epsilon_1\\    
 1 & \bar{\mathbb{A}}_{i,j}>\epsilon_1
\end{matrix}\right. ,
\label{eq_delete}
\end{equation}
where $(i, j)\in \mathcal{E}$ and $\epsilon_1 \in [-1, 1]$.
Subsequently, locality-based edge addition is implemented to incorporate potentially effective edges and further enhance the quality of the graph.
To reduce the search space for added edges, we adhere to the graph homophily principle and restrict the neighbor candidates of each target node to its $L$-hop neighbors in the original training graph.
Specifically, we first derive the $L$-hop edge set $\mathcal{E}_L$ by calculating ${\mathbf{A}^{L}}$ and then further reduce the candidate edges to $\mathcal{E}_L^r$ by selecting $r$ nearest neighbors for each target node.
Finally, we preserve the reliable edges whose reliability is larger than $\epsilon_2$:
\begin{equation}
{\mathbb{A}}_{i,j}=\left\{\begin{matrix}
 0 & \bar{\mathbb{A}}_{i,j}\le \epsilon_2\\    
 1 & \bar{\mathbb{A}}_{i,j}>\epsilon_2
\end{matrix}\right. ,
\label{eq_add}
\end{equation}
where $(i, j)\in \mathcal{E}_L^r$ and $\epsilon_2 \in [-1, 1]$. 
Consequently, the graph structure ${\mathbf{A}}$ is denoised to produce the structure $\mathbb{A}$.

{Notably, the dual-stage structure modification does not require gradient calculations and can be performed on CPUs, enabling flexible management by processing edges and nodes in separate batches, respectively. Furthermore, this module is adaptable to incorporate more advanced sampling strategies to enhance scalability, such as prioritizing low-degree nodes that are more vulnerable in the message-passing mechanism~\cite{li2022reliable}. Detailed complexity analysis is presented in Section D.}

\subsection{Optimization Objective and Alternating Optimization}
With the definition of graph structure modeling, the graph quality is primarily controlled by thresholds ($\epsilon_1$, $\epsilon_2$). 
A viable method to obtaining the optimal thresholds entails leveraging a pre-trained GNN model for gradient-based optimization.
However, training the GNN model on the noisy training graph inevitably results in the compromised performance, and optimizing the discrete graph structure by gradients complicates the optimization process. 

In light of these challenges, we employ the label propagation mechanism \cite{wang2021combining} to guide the optimization procedure and determine the optimal thresholds by grid search. 
This training-free approach notably simplifies the optimization by eliminating hyper-parameter tuning and back-propagation procedures. To achieve this, 50\% of the training nodes are randomly selected as support nodes, and their labels are propagated to predict the labels of the remaining nodes.\footnote{{{We assume a clean label set to investigate structural noise. For scenarios with label noise, our method is compatible with various off-the-shelf label enhancement methods to pre-process the labels, such as ensemble learning~\cite{lu2022ensemble}, self-supervised learning~\cite{tan2021co}, and large language models~\cite{yuan2024hide}.}}} 
Specifically, we construct the propagation label matrix $\bar{\mathbf{Y}}$ by retaining the one-hot labels of support nodes and marking the remaining labels as $\mathbf{0}$.
Then, ${\bf A}$ is optimized based on a selected threshold candidate and $\bar{\mathbf{Y}}$ is propagated through the optimized graph structure ${\mathbb{A}}$ as:
\begin{equation}
\begin{aligned}
&\hat{\mathbf{Y}}^{(k)} = \alpha{\hat{\mathbb{A}}}\hat{\mathbf{Y}}^{(k-1)}+(1-\alpha)\hat{\mathbf{Y}}^{(0)},
\end{aligned}
\label{eq_LP}
\end{equation}
where $\hat{\mathbf{Y}}^{(0)}=\bar{\mathbf{Y}}$, ${\hat{\mathbb{A}}}$ is the normalized adjacency matrix, and $\alpha$ is the teleport probability \cite{page1999pagerank}. $\hat{\mathbf{Y}}^{(k)}$ is the propagated label at iteration $k$. Finally, the optimization objective is:
\begin{equation}
\begin{aligned}
&\arg\max_{\epsilon_1, \epsilon_2} \sum_{i=1}^{N}  \mathbbm{1}   \{{\mathbf{Y}_i}=\hat{\mathbf{Y}}_i^{(l)}\},
\end{aligned}
\label{eq_loss}
\end{equation}
where  $\mathbbm{1}\{\cdot\}$ is the indicator function, and the labels are propagated for $l$ iterations.
Our objective is to identify the optimal thresholds ($\epsilon_1$, $\epsilon_2$) that yield the highest accuracy on the training set.

As per our design of RobGC, the condensation procedure and reliable graph structure learning are heavily entangled and can thus interfere with each other during training. 
Therefore, we propose to optimize $\mathcal{S}$ and ${\bf A}$ in an alternating manner, which is done by optimize $\mathcal{S}$ or ${\bf A}$ with the other side fixed. The details of training procedure is provided in Algorithm \ref{algorithm}. 
Without loss of generality, we use GCond as an example and any other GC methods can also be integrated into this paradigm. 
To mitigate the impact of noisy structures on the condensed graph, a warm-up stage (line 3) is designed to optimize $\mathbf{A}$ based solely on node features ${\bf X}$ with Eq. (\ref{eq_cosine2}).

\subsection{Test-Time Denoising}
Based on our earlier discussions, the noisy structure in the test graph can significantly impair prediction results, even when models are trained on a clean condensed graph. 
To efficiently handle the noise in the test graph structure under inductive scenario, we implement the test-time denoising before inference as shown in Fig. \ref{fig_main}.
Specifically, as inductive nodes are integrated into the training graph, the test graph $\mathbf{A}_{test}$ is optimized according to the noise-free condensed graph and thresholds ($\epsilon_1$, $\epsilon_2$) derived from the condensation stage, as specified in Eq. (\ref{eq_cosine1})-(\ref{eq_add}). 
Following this optimization, GNN models are then applied to the refined structure $\mathbb{A}_{test}$ for inference.

\subsection{Complexity Analysis}
The time complexity of optimizing the graph structure encompasses three components: the calculation of the correlation matrix, correlation matrix propagation, and the computation of edge reliability. 
Specifically, the time complexity for calculating the correlation matrix and propagation are  $\mathcal{O}(dN'N)$ and $\mathcal{O}(KN'^2N)$ respectively, where $d$ indicates the node dimension and $K$ is the number of propagation.
Contributing to the dual-stage structure optimization and the sparse structure of graph, the complexity of computing edge reliability is significantly reduced from $\mathcal{O}((d+KN')N^2)$ to $\mathcal{O}((d+KN')|\mathcal{E}_L|)$.
Consequently, the overall time complexity is formulated as $\mathcal{O}\left(\left(d+KN'\right)\left(N'N+|\mathcal{E}_L|\right)\right)$. In terms of space complexity, RobGC necessitates an additional $\mathcal{O}\left((K+1)N'N\right)$ and $\mathcal{O}(|\mathcal{E}_L|)$ over the base GC method for storing the correlation matrices and edge reliability.

\begin{algorithm}[t]
\SetAlgoVlined
\textbf{Input:} Training graph $\mathcal{T}=\{{\bf A}, {\bf X}\}$, initialised $\mathcal{S}$.\\
\textbf{Output:} Condensed graph $\mathcal{S}$, thresholds ($\epsilon_1$, $\epsilon_2$).\\
Warm-up to get the optimized graph structure ${\mathbb A}$.\\
\For{$t = 1,\ldots,T$}{
{\color{gray}{$\rhd$ Optimize condensed graph ${\mathcal{S}}$}}\\
Compute $\mathcal{L}_{cond}$ according to ${\mathbb A}$ with Eq. (\ref{eq_gmloss}).\\
Update ${\mathcal S}$ according to $\mathcal{L}_{cond}$.\\
{\color{gray}{$\rhd$ Optimize graph structure ${\mathbf{A}}$}}\\
\If{$t\%\tau = 0$}{
Calculate reliability $\bar{\mathbb{A}}$ for $\mathcal{E}$ and $\mathcal{E}_L$ by Eq. (\ref{eq_cosine2}).\\
\For{$\epsilon_1$ = $\min({\bar{\mathbb{A}}})$ to $\max({\bar{\mathbb{A}}})$ by $c$}{
\For{$\epsilon_2$ = $\min({\bar{\mathbb{A}}})$ to $\max({\bar{\mathbb{A}}})$ by $c$}{
Optimize ${\mathbf{A}}$ with Eq. (\ref{eq_delete}) and (\ref{eq_add}).\\
Propagate label with Eq. (\ref{eq_LP}).\\
}}
Compute the optimal thresholds ($\epsilon_1$, $\epsilon_2$).\\
Calculate the optimized graph structure ${\mathbb A}$.\\
}}
\textbf{Return:} Condensed graph $\mathcal{S}$ and thresholds ($\epsilon_1$, $\epsilon_2$).\\
\caption{Alternating Optimization of RobGC}
\label{algorithm}
\end{algorithm}

\section{Experiments}

We design comprehensive experiments to assess the effectiveness of RobGC and explore the following research questions:
\textbf{Q1}: When combined with different GC methods, can RobGC maintain robust performance under varying levels of noise?
\textbf{Q2}: Can RobGC effectively defend against various types of noise?
\textbf{Q3}: How does RobGC perform under different condensation ratios?
\textbf{Q4}: Does the condensed graph produced by RobGC exhibit good generalization across different GNN architectures?
\textbf{Q5}: How does the computational efficiency of RobGC?
\textbf{Q6}: How do the distinct components within RobGC influence its overall performance?
\textbf{Q7}: How does the optimized graph differ from the noisy training graph?
\textbf{Q8}: How do the different hyper-parameters affect the RobGC?
 {{\textbf{Q9}: Can Robust GNNs benefit from RobGC?}}

\subsection{Experimental Settings}
\label{exp_set}

\noindent\textbf{Datasets}. 
{We evaluate the methods on 8 datasets \cite{jin2022graph, liu_graph_2023} under our inductive setting, as summarized in Table \ref{tab:data}.}

\begin{table}[t]
\renewcommand{\arraystretch}{1.2}
\setlength{\abovecaptionskip}{0.1cm}
\centering
\caption{{{The properties of datasets.}}}
\label{tab:data}
\resizebox{\linewidth}{!}
{\begin{tabular}{l|rrrrrr}
\hline
Dataset    & \#Nodes & \#Edges   & \#Train/Val/Test nodes & Train/Val/Test graph size \\ \hline
Cora       & 2708   & 5429               & 1208/500/1000 & 1208/1708/2208 \\ 
Citeseer   & 3327   & 4732              & 1827/500/1000 & 1827/2327/2827  \\ 
Pubmed     & 19717   & 44338             & 18217/500/1000 & 18217/18717/19217 \\ 
Ogbn-arxiv & 169343  & 1166243         & 90941/29799/48603 & 90941/120740/139544    \\ 

Flickr & 89250  &  899756         & 44625/22312/22313 & 44625/66937/66938    \\ 
Reddit & 232965  & 57307946         & 153932/23699/55334 & 153932/177631/209266    \\ 

Squirrel &  5201  & 396846         & 3701/500/1000 & 3701/4201/4701    \\ 
Gamers &  168114  &  13595114         &  84056/42028/42030 & 84056/126084/126086    \\

\hline
\end{tabular}}
\end{table}

\begin{table*}[th]
\renewcommand{\arraystretch}{1.2}
\setlength{\abovecaptionskip}{0.1cm}
\centering
\caption{{{
The performance comparison of denoising methods under \textit{random noise} conditions. 
Whole denotes that GNNs are trained on the noisy training graph directly and evaluated on the noisy test graph.
Plain indicates that GNNs are trained on the condensed graph without denoising and evaluated on the noisy test graph.}}
}
\label{tab_acc}
\resizebox{\textwidth}{!}{
\begin{tabular}{cl|cccc|cccc|cccc|cccc}
\hline
\multicolumn{2}{l|}{Dataset ($r$)}                             & \multicolumn{4}{c|}{Cora (5.80\%)}                                            & \multicolumn{4}{c|}{Citeseer (3.30\%)}                                        & \multicolumn{4}{c|}{Pubmed (0.17\%)}                                          & \multicolumn{4}{c}{Ogbn-arxiv (0.50\%)}                                       \\ \hline
\multicolumn{2}{l|}{Noise level (\%)}                          & 0                 & 20                & 40                & 100               & 0                 & 20                & 40                & 100               & 0                 & 20                & 40                & 100               & 0                 & 20                & 40                & 100               \\ \hline
\multicolumn{2}{l|}{Whole Dataset}                             & 86.8\tiny{±0.3}          & 82.9\tiny{±0.4}          & 78.8\tiny{±0.3}          & 70.9\tiny{±1.1}          & 77.7\tiny{±0.4}          & 75.2\tiny{±0.2}          & 72.8\tiny{±0.1}          & 69.2\tiny{±0.3}          & 88.0\tiny{±0.1}          & 86.0\tiny{±0.2}          & 81.9\tiny{±0.3}          & 77.5\tiny{±0.2}          & 69.9\tiny{±0.1}          & 66.4\tiny{±0.1}          & 64.6\tiny{±0.3}          & 61.2\tiny{±0.1}          \\ \hline
\multicolumn{1}{c|}{\multirow{6}{*}{GCond}} & Plain            & 80.5\tiny{±0.1}          & 77.4\tiny{±0.2}          & 74.0\tiny{±0.3}          & 64.7\tiny{±0.8}          & 75.5\tiny{±0.3}          & 70.7\tiny{±1.2}          & 70.1\tiny{±0.2}          & 66.2\tiny{±0.1}          & 81.4\tiny{±0.1}          & 78.1\tiny{±0.2}          & 76.2\tiny{±0.1}          & 69.8\tiny{±0.3}          & 59.3\tiny{±0.1}          & 53.7\tiny{±0.8}          & 52.2\tiny{±0.5}          & 46.4\tiny{±0.5}          \\
\multicolumn{1}{c|}{}                       & Jaccard          & 79.9\tiny{±1.4}          & 75.9\tiny{±0.1}          & 74.2\tiny{±0.6}          & 67.5\tiny{±0.5}          & 74.0\tiny{±0.2}          & 71.8\tiny{±0.5}          & 71.6\tiny{±0.3}          & 68.0\tiny{±0.3}          & 81.0\tiny{±0.4}          & 78.3\tiny{±0.2}          & 75.8\tiny{±0.2}          & 73.7\tiny{±0.6}          & 58.2\tiny{±0.2}          & 54.1\tiny{±0.4}          & 53.1\tiny{±0.2}          & 47.8\tiny{±0.2}          \\
\multicolumn{1}{c|}{}                       & SVD              & 77.8\tiny{±0.1}          & 76.9\tiny{±0.3}          & 74.3\tiny{±0.3}          & 68.5\tiny{±0.5}          & 74.3\tiny{±0.2}          & 72.0\tiny{±0.2}          & 70.4\tiny{±0.3}          & 68.3\tiny{±0.2}          & 80.3\tiny{±0.6}          & 78.6\tiny{±0.2}          & 78.2\tiny{±0.1}          & 75.5\tiny{±0.7}          & 58.3\tiny{±0.2}          & 53.5\tiny{±0.4}          & 53.1\tiny{±0.5}          & 48.0\tiny{±0.8}          \\
\multicolumn{1}{c|}{}                       & kNN              & 78.4\tiny{±0.2}          & 77.8\tiny{±0.4}          & 74.6\tiny{±0.5}          & 65.7\tiny{±0.7}          & 74.3\tiny{±0.1}          & 71.6\tiny{±0.7}          & 71.2\tiny{±1.0}          & 67.3\tiny{±0.2}          & 80.6\tiny{±0.2}          & 80.4\tiny{±0.2}          & 78.8\tiny{±0.3}          & 73.6\tiny{±0.8}          & 59.0\tiny{±0.7}          & 54.2\tiny{±0.7}          & 53.0\tiny{±0.8}          & 47.1\tiny{±1.4}          \\
\multicolumn{1}{c|}{}                       & ${\rm STABLE_G}$ & 80.6\tiny{±0.2}          & 77.9\tiny{±0.3}          & 75.0\tiny{±0.2}          & 69.3\tiny{±0.2}          & 75.5\tiny{±0.1}          & 72.8\tiny{±0.4}          & 72.6\tiny{±0.1}          & 70.0\tiny{±0.3}          & 81.3\tiny{±0.4}          & 79.1\tiny{±0.2}          & 78.8\tiny{±0.2}          & 75.7\tiny{±0.6}          & 59.3\tiny{±0.2}          & 54.1\tiny{±0.2}          & 53.3\tiny{±0.3}          & 48.5\tiny{±0.1}          \\
\multicolumn{1}{c|}{}                       & Ours             & \textbf{80.9\tiny{±0.5}} & \textbf{78.9\tiny{±0.7}} & \textbf{76.3\tiny{±0.3}} & \textbf{70.6\tiny{±0.2}} & \textbf{75.9\tiny{±0.4}} & \textbf{75.5\tiny{±0.1}} & \textbf{75.1\tiny{±0.2}} & \textbf{72.9\tiny{±0.1}} & \textbf{81.5\tiny{±0.1}} & \textbf{80.8\tiny{±0.1}} & \textbf{80.7\tiny{±0.7}} & \textbf{76.8\tiny{±0.3}} & \textbf{59.6\tiny{±0.2}} & \textbf{55.0\tiny{±0.5}} & \textbf{54.0\tiny{±0.3}} & \textbf{49.7\tiny{±0.3}} \\ \hline
\multicolumn{1}{c|}{\multirow{6}{*}{GCDM}}  & Plain            & 77.7\tiny{±0.7}          & 68.0\tiny{±1.1}          & 63.5\tiny{±0.9}          & 53.9\tiny{±1.3}          & 74.7\tiny{±0.1}          & 70.9\tiny{±0.4}          & 69.8\tiny{±0.2}          & 60.5\tiny{±0.4}          & 83.4\tiny{±0.4}          & 76.7\tiny{±0.6}          & 74.9\tiny{±0.3}          & 70.7\tiny{±0.2}          & 57.8\tiny{±0.3}          & 52.2\tiny{±0.4}          & 52.0\tiny{±0.2}          & 46.2\tiny{±0.5}          \\
\multicolumn{1}{c|}{}                       & Jaccard          & 77.6\tiny{±0.5}          & 72.1\tiny{±0.2}          & 68.5\tiny{±0.2}          & 61.1\tiny{±0.8}          & 74.5\tiny{±0.1}          & 72.7\tiny{±0.2}          & 70.6\tiny{±0.4}          & 64.7\tiny{±0.4}          & 82.3\tiny{±0.2}          & 78.4\tiny{±0.1}          & 75.9\tiny{±0.1}          & 72.6\tiny{±0.1}          & 58.7\tiny{±0.2}          & 55.0\tiny{±0.3}          & 53.3\tiny{±0.5}          & 48.3\tiny{±0.8}          \\
\multicolumn{1}{c|}{}                       & SVD              & 76.4\tiny{±0.3}          & 73.3\tiny{±0.2}          & 71.5\tiny{±0.2}          & 57.6\tiny{±2.7}          & 74.6\tiny{±0.3}          & 71.2\tiny{±0.2}          & 70.4\tiny{±0.5}          & 62.7\tiny{±0.4}          & 79.6\tiny{±0.6}          & 78.0\tiny{±0.3}          & 77.6\tiny{±0.8}          & 76.4\tiny{±0.1}          & 57.5\tiny{±0.5}          & 53.2\tiny{±0.4}          & 52.4\tiny{±0.2}          & 49.2\tiny{±0.2}          \\
\multicolumn{1}{c|}{}                       & kNN              & 76.2\tiny{±0.5}          & 68.6\tiny{±0.4}          & 64.9\tiny{±1.3}          & 55.3\tiny{±1.2}          & 74.1\tiny{±0.2}          & 72.2\tiny{±0.5}          & 70.0\tiny{±0.2}          & 61.3\tiny{±0.2}          & 82.2\tiny{±0.4}          & 79.1\tiny{±0.5}          & 76.6\tiny{±0.6}          & 71.2\tiny{±0.2}          & 57.4\tiny{±3.5}          & 52.3\tiny{±0.2}          & 52.8\tiny{±0.9}          & 48.3\tiny{±0.4}          \\
\multicolumn{1}{c|}{}                       & ${\rm STABLE_G}$ & 77.8\tiny{±0.4}          & 73.5\tiny{±0.2}          & 71.7\tiny{±0.2}          & 62.0\tiny{±0.7}          & 74.8\tiny{±0.2}          & 72.9\tiny{±0.3}          & 71.2\tiny{±0.2}          & 67.7\tiny{±0.7}          & 83.4\tiny{±0.3}          & 78.6\tiny{±0.4}          & 77.2\tiny{±0.4}          & 75.6\tiny{±0.3}          & 58.8\tiny{±0.3}          & 54.8\tiny{±0.3}          & 53.4\tiny{±0.1}          & 49.0\tiny{±0.3}          \\
\multicolumn{1}{c|}{}                       & Ours             & \textbf{77.8\tiny{±0.3}} & \textbf{74.1\tiny{±0.2}} & \textbf{72.4\tiny{±0.2}} & \textbf{63.8\tiny{±0.6}} & \textbf{74.9\tiny{±0.1}} & \textbf{73.7\tiny{±0.1}} & \textbf{71.9\tiny{±0.1}} & \textbf{70.2\tiny{±0.1}} & \textbf{83.6\tiny{±0.2}} & \textbf{79.2\tiny{±0.1}} & \textbf{78.5\tiny{±0.2}} & \textbf{77.3\tiny{±0.4}} & \textbf{59.2\tiny{±0.1}} & \textbf{55.4\tiny{±0.2}} & \textbf{53.9\tiny{±0.3}} & \textbf{49.3\tiny{±0.4}} \\ \hline
\multicolumn{1}{c|}{\multirow{6}{*}{GDEM}}  & Plain            & 80.9\tiny{±0.3}          & 78.9\tiny{±0.3}          & 75.2\tiny{±0.4}          & 65.1\tiny{±0.5}          & 75.9\tiny{±0.6}          & 72.1\tiny{±0.4}          & 71.5\tiny{±0.4}          & 67.5\tiny{±0.5}          & 82.0\tiny{±0.4}          & 78.8\tiny{±0.4}          & 77.4\tiny{±0.3}          & 71.2\tiny{±0.2}          & 59.9\tiny{±0.3}          & 55.9\tiny{±0.4}          & 53.6\tiny{±0.3}          & 49.1\tiny{±0.3}          \\
\multicolumn{1}{c|}{}                       & Jaccard          & 81.1\tiny{±0.3}          & 80.8\tiny{±0.2}          & 76.3\tiny{±0.5}          & 68.7\tiny{±0.4}          & 76.2\tiny{±0.2}          & 72.6\tiny{±0.2}          & 71.8\tiny{±0.1}          & 68.3\tiny{±0.6}          & 82.2\tiny{±0.3}          & 80.3\tiny{±0.1}          & 78.2\tiny{±0.2}          & 73.0\tiny{±0.2}          & 58.8\tiny{±0.8}          & 56.5\tiny{±0.1}          & 54.0\tiny{±0.3}          & 49.8\tiny{±0.2}          \\
\multicolumn{1}{c|}{}                       & SVD              & 79.6\tiny{±0.3}          & 79.3\tiny{±0.9}          & 76.1\tiny{±0.3}          & 69.5\tiny{±0.3}          & 75.4\tiny{±0.4}          & 72.8\tiny{±0.1}          & 71.7\tiny{±0.1}          & 68.4\tiny{±0.2}          & 81.2\tiny{±0.3}          & 80.8\tiny{±0.4}          & 79.6\tiny{±0.1}          & 76.0\tiny{±0.2}          & 59.5\tiny{±0.1}          & 57.1\tiny{±0.3}          & 54.3\tiny{±0.4}          & 49.7\tiny{±0.8}          \\
\multicolumn{1}{c|}{}                       & kNN              & 81.0\tiny{±0.5}          & 80.7\tiny{±0.3}          & 76.2\tiny{±0.2}          & 69.7\tiny{±0.4}          & 76.1\tiny{±0.4}          & 72.4\tiny{±0.3}          & 71.5\tiny{±0.3}          & 68.2\tiny{±0.2}          & 79.7\tiny{±0.7}          & 79.1\tiny{±0.2}          & 78.0\tiny{±0.4}          & 74.9\tiny{±1.6}          & 59.3\tiny{±0.5}          & 56.8\tiny{±0.7}          & 54.8\tiny{±0.6}          & 49.3\tiny{±1.9}          \\
\multicolumn{1}{c|}{}                       & ${\rm STABLE_G}$ & 81.3\tiny{±0.2}          & 80.9\tiny{±0.2}          & 77.4\tiny{±0.3}          & 68.5\tiny{±0.3}          & 76.2\tiny{±0.2}          & 73.2\tiny{±0.3}          & 72.0\tiny{±0.3}          & 68.5\tiny{±0.6}          & 82.1\tiny{±0.4}          & 80.5\tiny{±0.1}          & 80.1\tiny{±0.2}          & 77.4\tiny{±0.5}          & 60.1\tiny{±0.9}          & 56.9\tiny{±0.7}          & 55.1\tiny{±0.4}          & 50.2\tiny{±0.3}          \\
\multicolumn{1}{c|}{}                       & Ours             & \textbf{82.5\tiny{±0.2}} & \textbf{81.8\tiny{±0.4}} & \textbf{79.9\tiny{±0.5}} & \textbf{70.9\tiny{±0.2}} & \textbf{76.3\tiny{±0.2}} & \textbf{76.0\tiny{±0.3}} & \textbf{75.4\tiny{±0.1}} & \textbf{73.2\tiny{±0.2}} & \textbf{82.2\tiny{±0.2}} & \textbf{81.8\tiny{±0.2}} & \textbf{80.8\tiny{±0.1}} & \textbf{78.9\tiny{±0.2}} & \textbf{60.5\tiny{±0.1}} & \textbf{58.2\tiny{±0.2}} & \textbf{55.8\tiny{±0.3}} & \textbf{52.1\tiny{±0.3}} \\ \hline
\multicolumn{2}{l|}{Dataset ($r$)}                             & \multicolumn{4}{c|}{Flickr (0.5\%)}                                           & \multicolumn{4}{c|}{Reddit (0.1\%)}                                           & \multicolumn{4}{c|}{Squirrel (1.7\%)}                                         & \multicolumn{4}{c}{Gamers (0.5\%)}                                            \\ \hline
\multicolumn{2}{l|}{Noise level (\%)}                          & 0                 & 20                & 40                & 100               & 0                 & 20                & 40                & 100               & 0                 & 20                & 40                & 100               & 0                 & 20                & 40                & 100               \\ \hline
\multicolumn{2}{l|}{Whole Dataset}                             & 50.9\tiny{±0.1}          & 50.2\tiny{±0.1}          & 49.7\tiny{±0.1}          & 48.9\tiny{±0.1}          & 95.3\tiny{±0.1}          & 93.9\tiny{±0.0}          & 93.3\tiny{±0.0}          & 92.1\tiny{±0.0}          & 44.8\tiny{±0.5}          & 35.3\tiny{±0.0}          & 27.5\tiny{±0.1}          & 24.3\tiny{±0.2}          & 62.1\tiny{±0.0}          & 61.9\tiny{±0.0}          & 61.4\tiny{±0.1}          & 60.8\tiny{±0.0}          \\ \hline
\multicolumn{1}{c|}{\multirow{6}{*}{GCond}} & Plain            & 47.9\tiny{±0.1}          & 47.5\tiny{±0.1}          & 47.5\tiny{±0.1}          & 47.0\tiny{±0.0}          & 90.7\tiny{±0.7}          & 89.5\tiny{±0.9}          & 88.7\tiny{±0.6}          & 87.0\tiny{±0.7}          & 36.6\tiny{±1.5}          & 27.0\tiny{±1.3}          & 22.4\tiny{±1.6}          & 21.8\tiny{±1.5}          & 57.6\tiny{±0.0}          & 57.4\tiny{±0.0}          & 56.8\tiny{±0.5}          & 56.1\tiny{±0.0}          \\
\multicolumn{1}{c|}{}                       & Jaccard          & 47.7\tiny{±0.2}          & 47.7\tiny{±0.5}          & 47.6\tiny{±0.1}          & 47.1\tiny{±0.4}          & 90.8\tiny{±0.5}          & 89.7\tiny{±0.5}          & 88.9\tiny{±0.5}          & 87.3\tiny{±0.5}          & 36.9\tiny{±1.2}          & 27.3\tiny{±1.0}          & 24.1\tiny{±1.4}          & 22.9\tiny{±1.9}          & 57.7\tiny{±0.3}          & 57.6\tiny{±0.1}          & 57.0\tiny{±0.3}          & 56.4\tiny{±0.1}          \\
\multicolumn{1}{c|}{}                       & SVD              & 47.8\tiny{±0.4}          & 47.6\tiny{±0.1}          & 47.6\tiny{±0.5}          & 47.2\tiny{±0.4}          & 90.9\tiny{±0.8}          & 89.7\tiny{±0.6}          & 89.4\tiny{±0.6}          & 87.5\tiny{±0.6}          & 36.8\tiny{±1.0}          & 27.5\tiny{±1.3}          & 24.2\tiny{±1.5}          & 23.6\tiny{±1.6}          & 57.9\tiny{±0.1}          & 57.6\tiny{±0.2}          & 57.1\tiny{±0.2}          & 56.4\tiny{±0.2}          \\
\multicolumn{1}{c|}{}                       & kNN              & 48.1\tiny{±0.2}          & 47.8\tiny{±0.2}          & 47.7\tiny{±0.0}          & 47.4\tiny{±0.1}          & 90.7\tiny{±0.5}          & 89.9\tiny{±0.8}          & 89.3\tiny{±0.4}          & 87.8\tiny{±0.4}          & 36.6\tiny{±1.7}          & 27.8\tiny{±1.7}          & 24.4\tiny{±1.7}          & 23.2\tiny{±1.6}          & 57.8\tiny{±0.1}          & 57.8\tiny{±0.1}          & 57.2\tiny{±0.1}          & 56.7\tiny{±0.1}          \\
\multicolumn{1}{c|}{}                       & ${\rm STABLE_G}$ & 47.9\tiny{±0.2}          & 47.7\tiny{±0.1}          & 47.7\tiny{±0.3}          & 47.2\tiny{±0.3}          & 90.9\tiny{±0.4}          & 90.1\tiny{±0.5}          & 89.5\tiny{±0.6}          & 87.8\tiny{±0.5}          & 36.7\tiny{±1.4}          & 27.7\tiny{±1.5}          & 24.4\tiny{±1.4}          & 23.7\tiny{±1.4}          & 57.9\tiny{±0.1}          & 57.7\tiny{±0.1}          & 57.2\tiny{±0.2}          & 56.6\tiny{±0.1}          \\
\multicolumn{1}{c|}{}                       & Ours             & \textbf{48.4\tiny{±0.0}} & \textbf{48.3\tiny{±0.0}} & \textbf{48.2\tiny{±0.1}} & \textbf{48.0\tiny{±0.1}} & \textbf{91.1\tiny{±0.5}} & \textbf{90.7\tiny{±0.9}} & \textbf{90.5\tiny{±0.7}} & \textbf{89.3\tiny{±0.7}} & \textbf{37.2\tiny{±1.5}} & \textbf{28.3\tiny{±1.4}} & \textbf{25.0\tiny{±1.2}} & \textbf{24.7\tiny{±1.9}} & \textbf{58.4\tiny{±0.2}} & \textbf{58.4\tiny{±0.1}} & \textbf{57.9\tiny{±0.3}} & \textbf{57.8\tiny{±0.1}} \\ \hline
\multicolumn{1}{c|}{\multirow{6}{*}{GCDM}}  & Plain            & 47.7\tiny{±0.0}          & 47.2\tiny{±0.1}          & 47.1\tiny{±0.2}          & 46.1\tiny{±0.1}          & 90.8\tiny{±0.2}          & 89.8\tiny{±0.4}          & 89.1\tiny{±0.6}          & 87.1\tiny{±0.5}          & 36.5\tiny{±2.7}          & 26.8\tiny{±1.7}          & 22.3\tiny{±1.4}          & 21.5\tiny{±0.9}          & 58.6\tiny{±0.2}          & 58.1\tiny{±0.2}          & 57.8\tiny{±0.1}          & 57.0\tiny{±0.1}          \\
\multicolumn{1}{c|}{}                       & Jaccard          & 47.9\tiny{±0.1}          & 47.6\tiny{±0.0}          & 47.3\tiny{±0.2}          & 46.4\tiny{±0.1}          & 91.0\tiny{±0.3}          & 89.7\tiny{±0.5}          & 88.9\tiny{±0.5}          & 87.4\tiny{±0.1}          & 36.6\tiny{±2.2}          & 27.2\tiny{±1.2}          & 23.8\tiny{±1.2}          & 22.2\tiny{±1.2}          & 58.7\tiny{±0.1}          & 58.5\tiny{±0.1}          & 58.1\tiny{±0.1}          & 57.2\tiny{±0.2}          \\
\multicolumn{1}{c|}{}                       & SVD              & 47.7\tiny{±0.1}          & 47.4\tiny{±0.0}          & 47.3\tiny{±0.0}          & 46.3\tiny{±0.0}          & 90.9\tiny{±0.4}          & 89.7\tiny{±0.6}          & 89.4\tiny{±0.6}          & 87.4\tiny{±0.5}          & 36.4\tiny{±2.1}          & 27.6\tiny{±1.4}          & 23.9\tiny{±2.3}          & 23.2\tiny{±1.6}          & 58.7\tiny{±0.2}          & 58.6\tiny{±0.1}          & 58.3\tiny{±0.1}          & 57.6\tiny{±0.1}          \\
\multicolumn{1}{c|}{}                       & kNN              & 47.9\tiny{±0.0}          & 47.7\tiny{±0.3}          & 47.5\tiny{±0.2}          & 46.4\tiny{±0.2}          & 91.0\tiny{±0.2}          & 89.9\tiny{±0.8}          & 89.3\tiny{±0.4}          & 88.0\tiny{±0.3}          & 36.7\tiny{±1.4}          & 27.6\tiny{±1.7}          & 23.2\tiny{±1.4}          & 22.1\tiny{±0.9}          & 59.0\tiny{±0.3}          & 58.8\tiny{±0.1}          & 58.6\tiny{±0.1}          & 57.7\tiny{±0.1}          \\
\multicolumn{1}{c|}{}                       & ${\rm STABLE_G}$ & 47.8\tiny{±0.1}          & 47.6\tiny{±0.0}          & 47.6\tiny{±0.2}          & 46.4\tiny{±0.0}          & 91.1\tiny{±0.3}          & 90.1\tiny{±0.5}          & 89.5\tiny{±0.6}          & 87.8\tiny{±0.3}          & 36.8\tiny{±1.8}          & 27.8\tiny{±1.6}          & 24.1\tiny{±1.1}          & 23.3\tiny{±1.3}          & 58.8\tiny{±0.2}          & 58.7\tiny{±0.2}          & 58.5\tiny{±0.1}          & 57.7\tiny{±0.2}          \\
\multicolumn{1}{c|}{}                       & Ours             & \textbf{48.3\tiny{±0.4}} & \textbf{48.1\tiny{±0.3}} & \textbf{48.1\tiny{±0.2}} & \textbf{47.3\tiny{±0.4}} & \textbf{91.3\tiny{±0.5}} & \textbf{91.0\tiny{±0.6}} & \textbf{90.6\tiny{±0.5}} & \textbf{89.5\tiny{±0.7}} & \textbf{37.0\tiny{±1.6}} & \textbf{28.2\tiny{±1.4}} & \textbf{24.7\tiny{±1.8}} & \textbf{23.9\tiny{±0.7}} & \textbf{59.4\tiny{±0.1}} & \textbf{59.2\tiny{±0.1}} & \textbf{59.1\tiny{±0.0}} & \textbf{58.6\tiny{±0.1}} \\ \hline
\multicolumn{1}{c|}{\multirow{6}{*}{GDEM}}  & Plain            & 48.5\tiny{±0.1}          & 48.0\tiny{±0.1}          & 47.7\tiny{±0.1}          & 47.4\tiny{±0.1}          & 91.7\tiny{±0.4}          & 91.5\tiny{±0.3}          & 90.4\tiny{±0.3}          & 89.4\tiny{±0.2}          & 37.9\tiny{±2.2}          & 28.2\tiny{±1.4}          & 23.6\tiny{±1.7}          & 22.5\tiny{±1.2}          & 59.7\tiny{±0.1}          & 59.4\tiny{±0.0}          & 59.0\tiny{±0.5}          & 58.3\tiny{±0.0}          \\
\multicolumn{1}{c|}{}                       & Jaccard          & 48.4\tiny{±0.2}          & 48.1\tiny{±0.1}          & 47.9\tiny{±0.3}          & 47.5\tiny{±0.3}          & 91.9\tiny{±0.1}          & 91.7\tiny{±0.1}          & 90.7\tiny{±0.1}          & 89.6\tiny{±0.4}          & 38.1\tiny{±1.2}          & 28.6\tiny{±1.0}          & 25.0\tiny{±1.1}          & 23.1\tiny{±1.4}          & 59.7\tiny{±0.3}          & 59.5\tiny{±0.1}          & 59.2\tiny{±0.2}          & 58.4\tiny{±0.1}          \\
\multicolumn{1}{c|}{}                       & SVD              & 48.5\tiny{±0.1}          & 48.3\tiny{±0.0}          & 47.8\tiny{±0.1}          & 47.6\tiny{±0.0}          & 91.6\tiny{±0.2}          & 91.6\tiny{±0.3}          & 90.7\tiny{±0.3}          & 89.7\tiny{±0.1}          & 38.2\tiny{±1.7}          & 28.8\tiny{±1.7}          & 25.2\tiny{±1.3}          & 24.8\tiny{±1.7}          & 59.8\tiny{±0.3}          & 59.6\tiny{±0.1}          & 59.4\tiny{±0.1}          & 58.5\tiny{±0.2}          \\
\multicolumn{1}{c|}{}                       & kNN              & 48.6\tiny{±0.4}          & 48.2\tiny{±0.1}          & 47.9\tiny{±0.2}          & 47.6\tiny{±0.1}          & 91.8\tiny{±0.5}          & 91.7\tiny{±0.6}          & 90.8\tiny{±0.4}          & 89.9\tiny{±0.2}          & 38.1\tiny{±1.9}          & 29.0\tiny{±1.2}          & 25.2\tiny{±1.2}          & 24.9\tiny{±1.2}          & 59.9\tiny{±0.2}          & 59.8\tiny{±0.1}          & 59.6\tiny{±0.3}          & 58.6\tiny{±0.1}          \\
\multicolumn{1}{c|}{}                       & ${\rm STABLE_G}$ & 48.6\tiny{±0.2}          & 48.3\tiny{±0.2}          & 47.9\tiny{±0.2}          & 47.7\tiny{±0.0}          & 91.8\tiny{±0.3}          & 91.8\tiny{±0.1}          & 90.8\tiny{±0.3}          & 90.0\tiny{±0.3}          & 37.9\tiny{±1.0}          & 29.4\tiny{±1.6}          & 25.3\tiny{±1.1}          & 24.7\tiny{±1.6}          & 59.9\tiny{±0.1}          & 59.7\tiny{±0.2}          & 59.7\tiny{±0.1}          & 58.6\tiny{±0.2}          \\
\multicolumn{1}{c|}{}                       & Ours             & \textbf{48.9\tiny{±0.2}} & \textbf{48.7\tiny{±0.2}} & \textbf{48.6\tiny{±0.1}} & \textbf{48.5\tiny{±0.1}} & \textbf{92.4\tiny{±0.1}} & \textbf{92.3\tiny{±0.1}} & \textbf{91.7\tiny{±0.3}} & \textbf{91.1\tiny{±0.2}} & \textbf{38.5\tiny{±1.3}} & \textbf{29.8\tiny{±1.1}} & \textbf{25.7\tiny{±1.1}} & \textbf{25.5\tiny{±1.9}} & \textbf{60.3\tiny{±0.0}} & \textbf{60.2\tiny{±0.1}} & \textbf{60.2\tiny{±0.0}} & \textbf{59.5\tiny{±0.1}} \\ \hline
\end{tabular}}
    \end{table*}

\begin{table*}[t]
\renewcommand{\arraystretch}{1.1}
\setlength{\abovecaptionskip}{0.1cm}
\centering
\caption{{{
The comparison of methods under \textit{adversarial attack} conditions. Refer to Table \ref{tab_acc} for Noise level=0\%.}}
}
\label{tab_adversarial}
\resizebox{\textwidth}{!}{
\begin{tabular}{cl|ccc|ccc|ccc|ccc}
\hline
\multicolumn{2}{l|}{Dataset ($r$)}                             & \multicolumn{3}{c|}{Cora (5.80\%)}                        & \multicolumn{3}{c|}{Citeseer (3.30\%)}                    & \multicolumn{3}{c|}{Pubmed (0.17\%)}                      & \multicolumn{3}{c}{Ogbn-arxiv (0.50\%)}                   \\ \hline
\multicolumn{2}{l|}{Noise level (\%)}                          & 5                 & 10                & 25                & 5                 & 10                & 25                & 5                 & 10                & 25                & 5                 & 10                & 25                \\ \hline
\multicolumn{2}{l|}{Whole Dataset}                             & 84.8\tiny{±0.1}          & 82.7\tiny{±0.1}          & 79.1\tiny{±0.9}          & 76.2\tiny{±1.1}          & 75.7\tiny{±0.2}          & 71.6\tiny{±0.3}          & 83.6\tiny{±0.1}          & 82.8\tiny{±0.2}          & 77.6\tiny{±0.3}          & 65.8\tiny{±0.2}          & 64.3\tiny{±0.1}          & 62.2\tiny{±0.1}          \\ \hline
\multicolumn{1}{c|}{\multirow{6}{*}{GCond}} & Plain            & 76.8\tiny{±0.3}          & 74.8\tiny{±0.7}          & 68.2\tiny{±0.3}          & 73.3\tiny{±0.4}          & 71.6\tiny{±0.3}          & 64.4\tiny{±0.4}          & 77.6\tiny{±0.3}          & 75.3\tiny{±0.2}          & 69.2\tiny{±0.1}          & 54.4\tiny{±0.2}          & 51.9\tiny{±0.2}          & 45.4\tiny{±0.2}          \\
\multicolumn{1}{c|}{}                       & Jaccard          & 77.1\tiny{±0.6}          & 76.6\tiny{±0.1}          & 72.1\tiny{±0.3}          & 73.9\tiny{±0.3}          & 71.9\tiny{±0.5}          & 67.7\tiny{±0.2}          & 77.4\tiny{±0.2}          & 76.1\tiny{±0.1}          & 70.4\tiny{±0.2}          & 54.8\tiny{±0.4}          & 52.3\tiny{±0.2}          & 46.4\tiny{±0.4}          \\
\multicolumn{1}{c|}{}                       & SVD              & 77.3\tiny{±0.4}          & 75.8\tiny{±0.2}          & 70.5\tiny{±0.6}          & 74.4\tiny{±0.1}          & 73.7\tiny{±0.2}          & 69.4\tiny{±0.4}          & 80.1\tiny{±0.3}          & 79.9\tiny{±0.3}          & 78.3\tiny{±0.3}          & 54.5\tiny{±2.9}          & 52.6\tiny{±2.4}          & 47.8\tiny{±3.0}          \\
\multicolumn{1}{c|}{}                       & kNN              & 77.1\tiny{±0.9}          & 75.3\tiny{±0.5}          & 71.6\tiny{±1.4}          & 74.7\tiny{±0.1}          & 72.7\tiny{±0.2}          & 67.0\tiny{±0.1}          & 78.0\tiny{±0.2}          & 76.7\tiny{±0.3}          & 70.8\tiny{±0.4}          & 54.6\tiny{±0.8}          & 52.9\tiny{±0.9}          & 46.0\tiny{±0.9}          \\
\multicolumn{1}{c|}{}                       & ${\rm STABLE_G}$ & 78.0\tiny{±0.5}          & 76.7\tiny{±0.4}          & 72.3\tiny{±0.5}          & 74.6\tiny{±0.1}          & 73.9\tiny{±0.3}          & 70.9\tiny{±0.3}          & 78.8\tiny{±0.2}          & 77.7\tiny{±0.2}          & 76.2\tiny{±0.3}          & 54.7\tiny{±0.8}          & 52.7\tiny{±0.3}          & 47.7\tiny{±0.4}          \\
\multicolumn{1}{c|}{}                       & Ours             & \textbf{78.7\tiny{±1.0}} & \textbf{77.8\tiny{±0.4}} & \textbf{73.9\tiny{±0.2}} & \textbf{75.8\tiny{±0.4}} & \textbf{75.4\tiny{±0.1}} & \textbf{73.4\tiny{±0.5}} & \textbf{80.6\tiny{±0.3}} & \textbf{79.9\tiny{±0.3}} & \textbf{78.7\tiny{±0.4}} & \textbf{54.8\tiny{±0.6}} & \textbf{53.1\tiny{±0.3}} & \textbf{48.2\tiny{±0.4}} \\ \hline
\multicolumn{1}{c|}{\multirow{6}{*}{GCDM}}  & Plain            & 72.7\tiny{±0.4}          & 71.0\tiny{±0.3}          & 58.9\tiny{±0.6}          & 73.1\tiny{±0.1}          & 71.6\tiny{±0.1}          & 56.7\tiny{±0.3}          & 78.1\tiny{±1.0}          & 75.4\tiny{±0.4}          & 67.9\tiny{±0.4}          & 54.5\tiny{±0.3}          & 48.0\tiny{±0.9}          & 44.5\tiny{±0.3}          \\
\multicolumn{1}{c|}{}                       & Jaccard          & 74.7\tiny{±0.9}          & 73.1\tiny{±0.6}          & 71.3\tiny{±0.6}          & 73.6\tiny{±0.1}          & 72.5\tiny{±0.2}          & 69.7\tiny{±0.4}          & 78.8\tiny{±0.4}          & 76.5\tiny{±0.3}          & 69.9\tiny{±0.4}          & 55.0\tiny{±0.5}          & 50.8\tiny{±0.2}          & 48.5\tiny{±0.4}          \\
\multicolumn{1}{c|}{}                       & SVD              & 73.7\tiny{±0.3}          & 73.1\tiny{±0.5}          & 65.2\tiny{±0.2}          & 73.4\tiny{±0.1}          & 71.9\tiny{±0.2}          & 67.6\tiny{±0.1}          & 79.0\tiny{±0.5}          & 78.6\tiny{±0.1}          & 76.0\tiny{±1.2}          & 54.8\tiny{±0.7}          & 48.7\tiny{±0.7}          & 46.1\tiny{±1.6}          \\
\multicolumn{1}{c|}{}                       & kNN              & 73.0\tiny{±0.3}          & 72.2\tiny{±1.7}          & 66.9\tiny{±0.7}          & 73.7\tiny{±0.7}          & 71.9\tiny{±0.3}          & 63.2\tiny{±0.5}          & 78.8\tiny{±0.4}          & 76.9\tiny{±0.1}          & 69.8\tiny{±0.2}          & 55.0\tiny{±0.2}          & 49.8\tiny{±0.5}          & 47.2\tiny{±0.6}          \\
\multicolumn{1}{c|}{}                       & ${\rm STABLE_G}$ & 74.9\tiny{±0.4}          & 74.1\tiny{±0.7}          & 70.9\tiny{±0.6}          & 73.7\tiny{±0.5}          & 72.8\tiny{±0.4}          & 70.2\tiny{±0.4}          & 79.1\tiny{±0.3}          & 78.9\tiny{±0.2}          & 75.8\tiny{±0.3}          & 55.2\tiny{±0.2}          & 51.0\tiny{±0.4}          & 48.8\tiny{±0.4}          \\
\multicolumn{1}{c|}{}                       & Ours             & \textbf{76.7\tiny{±0.5}} & \textbf{75.4\tiny{±0.1}} & \textbf{71.6\tiny{±1.2}} & \textbf{73.9\tiny{±0.1}} & \textbf{73.4\tiny{±0.2}} & \textbf{71.8\tiny{±0.1}} & \textbf{79.4\tiny{±0.2}} & \textbf{79.2\tiny{±0.2}} & \textbf{77.1\tiny{±0.1}} & \textbf{55.7\tiny{±0.2}} & \textbf{53.8\tiny{±0.1}} & \textbf{50.7\tiny{±0.3}} \\ \hline
\multicolumn{1}{c|}{\multirow{6}{*}{GDEM}}  & Plain            & 77.2\tiny{±0.4}          & 76.1\tiny{±0.4}          & 68.7\tiny{±0.3}          & 74.4\tiny{±0.3}          & 72.4\tiny{±0.2}          & 65.7\tiny{±0.3}          & 78.1\tiny{±0.4}          & 76.0\tiny{±0.3}          & 70.4\tiny{±0.2}          & 55.6\tiny{±0.3}          & 52.8\tiny{±0.6}          & 47.2\tiny{±0.3}          \\
\multicolumn{1}{c|}{}                       & Jaccard          & 79.3\tiny{±0.5}          & 77.3\tiny{±0.3}          & 72.1\tiny{±0.1}          & 75.3\tiny{±0.2}          & 73.2\tiny{±0.3}          & 68.3\tiny{±0.4}          & 79.1\tiny{±0.3}          & 76.2\tiny{±0.2}          & 71.9\tiny{±0.2}          & 56.8\tiny{±0.5}          & 53.6\tiny{±0.5}          & 49.6\tiny{±0.6}          \\
\multicolumn{1}{c|}{}                       & SVD              & 78.3\tiny{±0.2}          & 76.7\tiny{±0.2}          & 70.1\tiny{±0.2}          & 76.1\tiny{±0.3}          & 73.3\tiny{±0.5}          & 69.6\tiny{±0.3}          & 79.7\tiny{±0.3}          & 79.5\tiny{±0.5}          & 78.1\tiny{±0.6}          & 56.6\tiny{±0.2}          & 54.0\tiny{±0.2}          & 49.6\tiny{±0.3}          \\
\multicolumn{1}{c|}{}                       & kNN              & 78.3\tiny{±0.3}          & 76.9\tiny{±0.4}          & 71.2\tiny{±0.3}          & 75.6\tiny{±0.4}          & 73.2\tiny{±0.4}          & 69.2\tiny{±0.4}          & 79.8\tiny{±0.3}          & 77.2\tiny{±0.5}          & 72.1\tiny{±0.3}          & 55.7\tiny{±0.6}          & 54.1\tiny{±0.3}          & 48.9\tiny{±0.4}          \\
\multicolumn{1}{c|}{}                       & ${\rm STABLE_G}$ & 79.1\tiny{±0.4}          & 77.3\tiny{±0.4}          & 71.6\tiny{±0.3}          & 76.5\tiny{±0.2}          & 74.2\tiny{±0.2}          & 71.0\tiny{±0.5}          & 80.5\tiny{±0.4}          & 78.0\tiny{±0.3}          & 76.5\tiny{±0.4}          & 56.7\tiny{±0.4}          & 54.9\tiny{±0.5}          & 50.5\tiny{±0.4}          \\
\multicolumn{1}{c|}{}                       & Ours             & \textbf{80.3\tiny{±0.4}} & \textbf{79.1\tiny{±0.7}} & \textbf{74.0\tiny{±0.4}} & \textbf{77.2\tiny{±0.3}} & \textbf{76.2\tiny{±0.3}} & \textbf{74.1\tiny{±0.4}} & \textbf{81.7\tiny{±0.3}} & \textbf{80.8\tiny{±0.3}} & \textbf{79.0\tiny{±0.3}} & \textbf{57.9\tiny{±0.3}} & \textbf{55.5\tiny{±0.3}} & \textbf{51.8\tiny{±0.3}} \\ \hline
\multicolumn{2}{l|}{Dataset ($r$)}                             & \multicolumn{3}{c|}{Flickr (0.5\%)}                       & \multicolumn{3}{c|}{Reddit (0.1\%)}                       & \multicolumn{3}{c|}{Squirrel (1.7\%)}                     & \multicolumn{3}{c}{Gamers (0.5\%)}                        \\ \hline
\multicolumn{2}{l|}{Noise level (\%)}                          & 5                 & 10                & 25                & 5                 & 10                & 25                & 5                 & 10                & 25                & 5                 & 10                & 25                \\ \hline
\multicolumn{2}{l|}{Whole Dataset}                             & 48.8\tiny{±0.1}          & 47.0\tiny{±0.1}          & 46.4\tiny{±0.1}          & 93.4\tiny{±0.0}          & 92.7\tiny{±0.1}          & 91.9\tiny{±0.0}          & 39.8\tiny{±0.6}          & 34.5\tiny{±0.7}          & 31.5\tiny{±0.4}          & 59.0\tiny{±0.1}          & 58.4\tiny{±0.1}          & 58.0\tiny{±0.1}          \\ \hline
\multicolumn{1}{c|}{\multirow{6}{*}{GCond}} & Plain            & 46.9\tiny{±0.2}          & 44.5\tiny{±0.0}          & 43.1\tiny{±0.1}          & 88.9\tiny{±0.4}          & 88.1\tiny{±0.5}          & 86.8\tiny{±0.2}          & 32.3\tiny{±1.8}          & 25.8\tiny{±1.8}          & 25.2\tiny{±1.6}          & 54.9\tiny{±0.1}          & 54.2\tiny{±0.2}          & 53.9\tiny{±0.1}          \\
\multicolumn{1}{c|}{}                       & Jaccard          & 46.9\tiny{±0.1}          & 44.6\tiny{±0.3}          & 43.2\tiny{±0.3}          & 89.1\tiny{±0.3}          & 88.3\tiny{±0.4}          & 87.0\tiny{±0.2}          & 32.4\tiny{±1.5}          & 26.0\tiny{±1.2}          & 25.8\tiny{±1.5}          & 55.2\tiny{±0.2}          & 54.6\tiny{±0.1}          & 54.2\tiny{±0.2}          \\
\multicolumn{1}{c|}{}                       & SVD              & 47.0\tiny{±0.0}          & 44.6\tiny{±0.1}          & 43.3\tiny{±0.0}          & 89.3\tiny{±0.4}          & 88.4\tiny{±0.4}          & 87.2\tiny{±0.3}          & 32.5\tiny{±1.1}          & 25.9\tiny{±1.3}          & 25.6\tiny{±1.2}          & 55.4\tiny{±0.1}          & 54.7\tiny{±0.1}          & 54.3\tiny{±0.0}          \\
\multicolumn{1}{c|}{}                       & kNN              & 47.1\tiny{±0.1}          & 44.7\tiny{±0.2}          & 43.2\tiny{±0.1}          & 89.4\tiny{±0.2}          & 88.7\tiny{±0.2}          & 87.4\tiny{±0.3}          & 32.5\tiny{±1.6}          & 26.1\tiny{±1.1}          & 25.9\tiny{±1.5}          & 55.5\tiny{±0.1}          & 54.9\tiny{±0.1}          & 54.5\tiny{±0.2}          \\
\multicolumn{1}{c|}{}                       & ${\rm STABLE_G}$ & 47.1\tiny{±0.2}          & 44.7\tiny{±0.2}          & 43.2\tiny{±0.0}          & 89.6\tiny{±0.4}          & 88.8\tiny{±0.3}          & 87.5\tiny{±0.2}          & 32.4\tiny{±1.2}          & 26.4\tiny{±1.3}          & 26.2\tiny{±1.4}          & 55.4\tiny{±0.1}          & 54.8\tiny{±0.2}          & 54.4\tiny{±0.1}          \\
\multicolumn{1}{c|}{}                       & Ours             & \textbf{47.8\tiny{±0.0}} & \textbf{45.5\tiny{±0.2}} & \textbf{44.1\tiny{±0.1}} & \textbf{90.5\tiny{±0.6}} & \textbf{89.6\tiny{±0.3}} & \textbf{88.7\tiny{±0.1}} & \textbf{32.9\tiny{±1.3}} & \textbf{26.9\tiny{±1.1}} & \textbf{26.9\tiny{±1.5}} & \textbf{56.0\tiny{±0.2}} & \textbf{55.7\tiny{±0.1}} & \textbf{55.4\tiny{±0.0}} \\ \hline
\multicolumn{1}{c|}{\multirow{6}{*}{GCDM}}  & Plain            & 46.0\tiny{±0.2}          & 44.3\tiny{±0.2}          & 42.7\tiny{±0.4}          & 89.2\tiny{±0.4}          & 88.4\tiny{±0.5}          & 87.0\tiny{±0.2}          & 32.0\tiny{±0.9}          & 25.5\tiny{±1.2}          & 25.0\tiny{±1.2}          & 54.6\tiny{±0.2}          & 54.0\tiny{±0.1}          & 53.8\tiny{±0.1}          \\
\multicolumn{1}{c|}{}                       & Jaccard          & 46.3\tiny{±0.1}          & 44.5\tiny{±0.0}          & 42.9\tiny{±0.2}          & 89.4\tiny{±0.3}          & 88.6\tiny{±0.2}          & 87.3\tiny{±0.2}          & 32.2\tiny{±1.2}          & 25.7\tiny{±1.0}          & 25.5\tiny{±1.3}          & 55.0\tiny{±0.2}          & 54.3\tiny{±0.1}          & 54.2\tiny{±0.1}          \\
\multicolumn{1}{c|}{}                       & SVD              & 46.1\tiny{±0.1}          & 44.4\tiny{±0.1}          & 43.1\tiny{±0.0}          & 89.6\tiny{±0.1}          & 88.7\tiny{±0.3}          & 87.6\tiny{±0.3}          & 32.4\tiny{±1.0}          & 25.6\tiny{±1.1}          & 25.2\tiny{±1.7}          & 55.1\tiny{±0.1}          & 54.4\tiny{±0.1}          & 54.2\tiny{±0.1}          \\
\multicolumn{1}{c|}{}                       & kNN              & 46.3\tiny{±0.0}          & 44.5\tiny{±0.2}          & 43.2\tiny{±0.2}          & 89.6\tiny{±0.3}          & 88.7\tiny{±0.2}          & 87.7\tiny{±0.3}          & 32.2\tiny{±1.1}          & 25.7\tiny{±1.2}          & 25.3\tiny{±1.2}          & 55.1\tiny{±0.3}          & 54.6\tiny{±0.1}          & 54.4\tiny{±0.3}          \\
\multicolumn{1}{c|}{}                       & ${\rm STABLE_G}$ & 46.4\tiny{±0.1}          & 44.6\tiny{±0.0}          & 43.2\tiny{±0.1}          & 89.8\tiny{±0.3}          & 88.9\tiny{±0.1}          & 87.7\tiny{±0.2}          & 32.5\tiny{±1.6}          & 26.0\tiny{±1.0}          & 25.6\tiny{±1.5}          & 55.3\tiny{±0.1}          & 54.6\tiny{±0.2}          & 54.5\tiny{±0.1}          \\
\multicolumn{1}{c|}{}                       & Ours             & \textbf{47.2\tiny{±0.1}} & \textbf{45.3\tiny{±0.1}} & \textbf{43.9\tiny{±0.2}} & \textbf{91.3\tiny{±0.5}} & \textbf{89.9\tiny{±0.3}} & \textbf{88.8\tiny{±0.1}} & \textbf{33.9\tiny{±1.0}} & \textbf{26.6\tiny{±1.1}} & \textbf{26.4\tiny{±1.6}} & \textbf{55.9\tiny{±0.0}} & \textbf{55.4\tiny{±0.1}} & \textbf{55.4\tiny{±0.0}} \\ \hline
\multicolumn{1}{c|}{\multirow{6}{*}{GDEM}}  & Plain            & 47.3\tiny{±0.1}          & 45.0\tiny{±0.1}          & 43.6\tiny{±0.2}          & 90.6\tiny{±0.3}          & 89.8\tiny{±0.5}          & 89.1\tiny{±0.2}          & 33.0\tiny{±0.9}          & 27.9\tiny{±0.8}          & 25.8\tiny{±1.1}          & 56.1\tiny{±0.1}          & 55.9\tiny{±0.2}          & 55.6\tiny{±0.1}          \\
\multicolumn{1}{c|}{}                       & Jaccard          & 47.4\tiny{±0.1}          & 45.2\tiny{±0.1}          & 43.7\tiny{±0.1}          & 90.9\tiny{±0.2}          & 90.0\tiny{±0.3}          & 89.3\tiny{±0.2}          & 33.1\tiny{±1.3}          & 28.3\tiny{±1.1}          & 26.5\tiny{±1.4}          & 56.2\tiny{±0.2}          & 56.0\tiny{±0.1}          & 55.9\tiny{±0.2}          \\
\multicolumn{1}{c|}{}                       & SVD              & 47.6\tiny{±0.2}          & 45.3\tiny{±0.0}          & 43.6\tiny{±0.1}          & 90.7\tiny{±0.1}          & 90.1\tiny{±0.3}          & 89.4\tiny{±0.3}          & 33.3\tiny{±1.2}          & 28.2\tiny{±1.3}          & 26.3\tiny{±1.3}          & 56.4\tiny{±0.1}          & 56.1\tiny{±0.2}          & 56.0\tiny{±0.1}          \\
\multicolumn{1}{c|}{}                       & kNN              & 47.4\tiny{±0.4}          & 45.2\tiny{±0.1}          & 43.7\tiny{±0.5}          & 91.1\tiny{±0.4}          & 90.3\tiny{±0.2}          & 89.7\tiny{±0.4}          & 33.4\tiny{±1.4}          & 28.4\tiny{±1.2}          & 26.6\tiny{±1.2}          & 56.5\tiny{±0.1}          & 56.2\tiny{±0.1}          & 56.1\tiny{±0.1}          \\
\multicolumn{1}{c|}{}                       & ${\rm STABLE_G}$ & 47.6\tiny{±0.2}          & 45.3\tiny{±0.2}          & 43.8\tiny{±0.0}          & 91.0\tiny{±0.2}          & 90.2\tiny{±0.3}          & 89.6\tiny{±0.1}          & 33.4\tiny{±1.7}          & 28.5\tiny{±1.2}          & 26.7\tiny{±1.1}          & 56.4\tiny{±0.1}          & 56.2\tiny{±0.2}          & 56.2\tiny{±0.1}          \\
\multicolumn{1}{c|}{}                       & Ours             & \textbf{48.3\tiny{±0.1}} & \textbf{46.1\tiny{±0.0}} & \textbf{44.5\tiny{±0.1}} & \textbf{91.9\tiny{±0.3}} & \textbf{91.3\tiny{±0.2}} & \textbf{90.1\tiny{±0.2}} & \textbf{33.9\tiny{±1.3}} & \textbf{29.1\tiny{±1.2}} & \textbf{27.4\tiny{±1.0}} & \textbf{57.2\tiny{±0.2}} & \textbf{57.0\tiny{±0.1}} & \textbf{56.9\tiny{±0.0}} \\ \hline
\end{tabular}}
\end{table*}

\noindent\textbf{Baselines and Evaluation Settings}.
\label{sec_eval}
We deploy our proposed method on three GC methods with different optimization strategies:
(1) GCond \cite{jin2022graph}: the first GC method that matches the model gradients derived from the training and condensed graph;
(2) GCDM \cite{zhao2023dataset}: an efficient GC method that minimizes the discrepancy between distributions of the training and condensed graphs.
(3) GDEM \cite{liu_graph_2023}: a generalized GC method that aligns the training and condensed graph's eigenvectors.

Our primary objective is to enhance graph quality rather than merely optimizing model performance. 
Consequently, we select four state-of-the-art (SOTA) denoising techniques capable of addressing noise in inductive test settings.
These methods are deployed with the aforementioned GC methods for comparison:
(1) Jaccard \cite{jin2020graph}: it computes the Jaccard similarity for every pair of connected nodes and discards edges whose similarity is smaller than a specific threshold.
(2) SVD \cite{jin2020graph}: it decomposes the noisy graph through Singular Value Decomposition (SVD) and adopts the low-rank approximation for the denoised graph to reduce the influence of potential high-rank noises.
(3) kNN \cite{jin2021node}: it identifies the k-nearest neighbors for each node based on their features and incorporates potentially effective edges into the original graph.
{{(4) ${\rm STABLE_G}$ \cite{li2022reliable}: it refines the graph structure according to the node embeddings obtained from self-supervised training.}}

Following \cite{jin2020graph}, we assess our method's resilience to both \textbf{random noise} and \textbf{adversarial attacks}. 
For random noise, it involves randomly adding fake edges to and removing real edges from the graph. Given the computational intensity of most adversarial attack methods, we utilize the SOTA scalable attack method, PR-BCD \cite{geisler2021robustness}, to execute global adversarial attacks. 
We evaluate each method across different noise levels, defined by the ratio of changed edges in the original graph \cite{jin2020graph}.
{We evaluate random noise at levels \{0\%, 20\%, 40\%, 100\%\} and adversarial attacks at \{5\%, 10\%, 25\%\}.}
 We operate under the premise of uniform noise levels in real-world systems, indicating both training and test graphs share the identical noise level.

For a fair comparison, we follow the GCond \cite{jin2022graph} {{and evaluate the node classification performance of models based on classification accuracy}}.
Across all GC methods compared, SGC \cite{wu2019simplifying} is utilized as the relay model and the downstream GNN is GCN~\cite{DBLP:conf/iclr/KipfW17}.
We condense the training graph with $N$ nodes into a condensed graph with $N'$ nodes, with the condensation ratio determined by $r=\frac{N'}{N}$. Following GCond, the number of condensed nodes per class are selected as \{5, 10, 20\} for the Cora, Citeseer, and Pubmed, respectively.

\begin{table*}[t]
    \renewcommand{\arraystretch}{1.2}
    \setlength{\abovecaptionskip}{0.1cm}
    \centering
    \caption{{{The performance comparison of denoise methods across different condensation ratios. Each method is equipped with GCond to condense the graph. The random noise is applied and noise level is 100\%.}} }
    \label{tab_condense_rate}
    \resizebox{0.7\textwidth}{!}{
    \begin{tabular}{l|r|cccccc|l}
    \hline
    Dataset                     & \multicolumn{1}{c|}{$r$} & \multicolumn{1}{c}{Plain} & \multicolumn{1}{c}{Jaccard} & \multicolumn{1}{c}{SVD} & \multicolumn{1}{c}{kNN} & ${\rm STABLE_G}$ & \multicolumn{1}{c|}{Ours} & \multicolumn{1}{c}{Whole} \\ \hline
    \multirow{3}{*}{Cora}       & 2.90\%                   & 63.7\tiny{±0.7}                  & 65.6\tiny{±0.2}                    & 67.0\tiny{±0.6}                & 64.2\tiny{±0.1}                & 67.1\tiny{±0.2}         & \textbf{70.3\tiny{±0.1}}         & \multirow{3}{*}{70.9\tiny{±1.1}} \\
                                & 5.80\%                   & 64.7\tiny{±0.8}                  & 67.5\tiny{±0.5}                    & 68.5\tiny{±0.5}                & 65.7\tiny{±0.7}                & 69.3\tiny{±0.2}         & \textbf{70.6\tiny{±0.2}}         &                           \\
                                & 11.60\%                  & 65.8\tiny{±0.3}                  & 67.8\tiny{±0.4}                    & 69.3\tiny{±0.9}                & 67.5\tiny{±0.5}                & 69.5\tiny{±0.4}         & \textbf{72.6\tiny{±0.4}}         &                           \\ \hline
    \multirow{3}{*}{Citeseer}   & 1.65\%                   & 66.0\tiny{±0.7}                  & 68.0\tiny{±0.1}                    & 68.2\tiny{±0.3}                & 66.1\tiny{±0.4}                & 69.2\tiny{±0.3}         & \textbf{71.9\tiny{±0.1}}         & \multirow{3}{*}{69.2\tiny{±0.3}} \\
                                & 3.30\%                   & 66.2\tiny{±0.1}                  & 68.0\tiny{±0.3}                    & 68.3\tiny{±0.2}                & 67.3\tiny{±0.2}                & 70.0\tiny{±0.3}         & \textbf{72.9\tiny{±0.1}}         &                           \\
                                & 6.60\%                   & 68.0\tiny{±0.5}                  & 69.0\tiny{±0.2}                    & 69.1\tiny{±0.2}                & 69.7\tiny{±0.9}                & 70.2\tiny{±0.5}         & \textbf{74.7\tiny{±0.1}}         &                           \\ \hline
    \multirow{3}{*}{Pubmed}     & 0.09\%                   & 68.6\tiny{±0.4}                  & 72.9\tiny{±0.7}                    & 74.7\tiny{±0.2}                & 72.7\tiny{±0.5}                & 74.9\tiny{±0.4}         & \textbf{76.2\tiny{±0.2}}         & \multirow{3}{*}{77.5\tiny{±0.2}} \\
                                & 0.17\%                   & 69.8\tiny{±0.3}                  & 73.7\tiny{±0.6}                    & 75.5\tiny{±0.7}                & 73.6\tiny{±0.8}                & 75.7\tiny{±0.6}         & \textbf{76.8\tiny{±0.3}}         &                           \\
                                & 0.33\%                   & 70.8\tiny{±0.3}                  & 73.7\tiny{±0.2}                    & 76.4\tiny{±0.1}                & 74.4\tiny{±0.2}                & 75.9\tiny{±0.5}         & \textbf{78.9\tiny{±0.1}}         &                           \\ \hline
    \multirow{3}{*}{Ogbn-arxiv} & 0.25\%                   & 45.1\tiny{±0.5}                  & 46.8\tiny{±0.2}                    & 46.0\tiny{±0.6}                & 47.0\tiny{±0.2}                & 47.2\tiny{±0.4}         & \textbf{47.9\tiny{±0.4}}         & \multirow{3}{*}{61.2\tiny{±0.1}} \\
                                & 0.50\%                   & 46.4\tiny{±0.5}                  & 47.8\tiny{±0.2}                    & 48.0\tiny{±0.8}                & 47.1\tiny{±0.4}                & 48.5\tiny{±0.1}         & \textbf{49.7\tiny{±0.3}}         &                           \\
                                & 1.00\%                   & 46.6\tiny{±0.2}                  & 48.3\tiny{±0.2}                    & 48.2\tiny{±0.6}                & 47.8\tiny{±0.4}                & 50.1\tiny{±0.5}         & \textbf{51.7\tiny{±0.4}}         &                           \\ \hline
    \end{tabular}}
\end{table*}

\noindent\textbf{Hyper-parameters and Implementation}. 
The hyper-parameters for GC methods are configured as described in their respective papers if clarified, while others are determined through grid search on the validation set.
For all datasets, a 2-layer graph convolution is employed, and the width of the hidden layer is set to 256. 
The learning rate for the condensation process is determined through a search over the set \{1e-2, 5e-3, 1e-3, 5e-4, 1e-4\}. 
{{
For our method, the periodic parameter $\tau$ is set as 50. The number of propagation for correlation matrix is set as $K=2$. The teleport probability is fixed at $\alpha=0.9$, and the iteration of label propagation is set as $l=10$. 
In the process of locality-based edge addition, we investigate up to 3-hop neighbors (i.e., $L$).
To optimize the grid search, the step parameter $c$ is dynamically adjusted by defining the number of search candidates $h$ as:
\begin{equation}
\begin{aligned}
 c=\frac{\max({\bar{\mathbb{A}}})-\min({\bar{\mathbb{A}}})}{h-1},
\end{aligned}
\end{equation}
{where $\bar{\mathbb{A}}$ measures the reliability for each edge, as defined in Eq. (\ref{eq_cosine2}).
Consequently, the grid search is controlled by varying $h$, which is set to 10 for the Ogbn-arxiv, Flickr, Reddit, and Gamers datasets and 20 for all other datasets.}

For denoising techniques, the Jaccard similarity threshold for removing dissimilar edges is chosen from \{0.01, 0.02, 0.03, 0.04, 0.05\}. For SVD, the reduced rank of the perturbed graph is tuned from \{10, 50, 100, 200\}. For kNN, the number of neighbors is selected from \{1, 2, 3, 4, 5\}.
}}

To mitigate randomness, each experiment is repeated 5 times, and the mean accuracy and standard deviation are reported.
The codes are written in Python 3.9 and the operating system is Ubuntu 16.0. We use Pytorch 1.12.1 on CUDA 12.0 to train models on GPU. All experiments are conducted on a machine with Intel(R) Xeon(R) CPUs (Gold 6326 @ 2.90GHz) and NVIDIA GeForce A40 GPUs.

\subsection{Denoising Performance (Q1\&Q2)}
We present the accuracy of various denoising methods against different levels of random noise and adversarial attacks in Table~\ref{tab_acc} and Table~\ref{tab_adversarial}, respectively. 
In these tables, \textbf{Whole Dataset} refers to GNNs trained on the noisy training graph and deployed on the noisy test graph. 
Notice that it suffers from intensive computational costs due to the large scale of the training graph.
\textbf{Plain} represents the GNN trained on the condensed graph without denoising and deployed on the noisy test graph. 
{{Notice that the performance of Whole Dataset and Plain differs from the results reported in the graph condensation papers \cite{jin2022graph, zhao2023dataset, liu_graph_2023} due to the distinct experimental setting of our study, which involves smaller training graphs and inductive test graphs, as shown in Table \ref{tab:data}.}}
As noise levels increase, the performances drop and the gap grows between Whole Dataset and Plain, particularly notable on the dataset Ogbn-arxiv. This trend verifies that noise in the training graph significantly degrades the quality of the condensed graph produced by various GC methods.

Regarding results against random noise, the performances of GC methods are successfully improved when equipped with denoising baselines. 
As noise levels increase, the benefits of enhanced graph quality become more evident.
Nonetheless, due to the lack of label guidance for structure optimization, these baselines inevitably introduce unreliable edges or destroy the potentially effective structures, leading to sub-optimal performance in some noise-free scenarios.
Moreover, denoising baselines exhibit varying effectiveness across datasets; for instance, SVD significantly outperforms Jaccard and kNN on the Pubmed dataset across all GC methods.
Our RobGC consistently outperforms other baselines across different GC methods and noise levels. 
With the guidance of label propagation, RobGC can outperform Plain in noise-free conditions.
Even at high noise levels, RobGC maintains superior performance than baselines by a larger margin.
Remarkably, with substantial condensation ratios, RobGC achieves better results than Whole Dataset on the Citeseer when equipped with GCond. 
This is attributed to the effectiveness of test-time denoising and enhanced test graph quality.
{Among various GC methods, GDEM achieves superior performance compared to GCond and GCDM. However, RobGC further enhances GDEM’s results, highlighting its strong competitiveness.}

The adversarial attack has a greater impact on classification performance compared to random noise at the same noise level. However, our method significantly improves condensed graph quality, standing out against other baselines.
These findings demonstrate the effectiveness of our proposed method in enhancing the quality of both the condensed graph and test data under various noise types.

\subsection{Performance Across Condensation Ratios (Q3)}

The condensation ratio plays a pivotal role in GC, directly influencing the quality of the condensed graph.
To understand how the condensation ratio affects denoising efficacy, we combine GCond with various denoising methods and evaluate their performance across different condensation ratios following GCond. As indicated in Table \ref{tab_condense_rate}, all denoising methods consistently improve performance across a range of condensation ratios, underscoring the importance of graph quality enhancement even at higher condensation levels. 
Among baselines, SVD achieves better performance, particularly on the Cora and Pubmed datasets. 
Our proposed method demonstrates superior performance over the other baselines across all condensation ratios. 
The advantage of our approach becomes apparent with increasing condensation ratios, where the improvements in condensed graph quality significantly boost denoising outcomes. 
Benefiting from the test-time denoising, the performance of our proposed method can even surpass the Whole on Citeseer and Pubmed.
Notably, as we compare performance across different condensation ratios, the improvement offered by our method becomes more pronounced with higher condensation ratios. This trend suggests that enhancements in the quality of the condensed graph play a crucial role, leading to improved denoising performance. The underlying rationale for this could be the alternating optimization process, which effectively leverages the improved graph quality to achieve better results.

\subsection{Generalizability for GNN Architectures (Q4)}
A critical attribute of GC is its ability to generalize across different GNN architectures, thereby allowing the condensed graph to serve as a versatile foundation for training multiple GNN models in downstream tasks.
To investigate this attribute, we assess the performance of different GNN models, including GCN~\cite{DBLP:conf/iclr/KipfW17}, SGC~\cite{wu2019simplifying}, GraphSAGE~\cite{hamilton2017inductive} and APPNP~\cite{klicpera2018predict}, using condensed graphs produced by GC methods combined with our proposed method.
These well-trained models are then deployed on the test graph after test-time denoising and results are presented in Table \ref{tab_crossarc}.
While performance varied among the condensation methods, all GNN models tested exhibited similar levels of effectiveness within each specific GC method. 
This consistency underscores the inherent generalizability of GC methods. Moreover, a significant factor contributing to this success is our proposed optimization strategy based on label propagation.  
This strategy eliminates the introduction of additional GNN models, therefore it not only simplifies the optimization process but also ensures broad architectural generalizability, highlighting the effectiveness of our method in maintaining high performance across various architectures.

\begin{table}[]
\renewcommand{\arraystretch}{1.2}
\centering
\setlength{\abovecaptionskip}{0.1cm}
\caption{{{Architecture generalizability of RobGC equipped with different graph condensation methods. Downstream GNNs are trained on condensed graphs for evaluation. The random noise is applied and the noise level is 100\%.}}}
\label{tab_crossarc}
\resizebox{\linewidth}{!}
{
\begin{tabular}{l|l|cccc}
\hline
\multirow{2}{*}{Dataset ($r$)}                                                 & \multirow{2}{*}{GC Method} & \multicolumn{4}{c}{Downstream GNNs}        \\ \cline{3-6} 
                                                                               &                            & GCN      & SGC      & GraphSAGE & APPNP    \\ \hline
\multirow{3}{*}{\begin{tabular}[c]{@{}l@{}}Cora \\ (5.80\%)\end{tabular}}      & GCond+Ours                 & 70.6\tiny{±0.2} & 72.7\tiny{±3.5} & 71.9\tiny{±0.3}  & 72.9\tiny{±0.6} \\
                                                                               & GCDM+Ours                  & 63.8\tiny{±0.6} & 64.3\tiny{±0.1} & 62.3\tiny{±0.6}  & 62.4\tiny{±0.1} \\
                                                                               & GDEM+Ours                  & 70.9\tiny{±0.2} & 72.9\tiny{±0.2} & 72.1\tiny{±0.3}  & 72.3\tiny{±0.4} \\ \hline
\multirow{3}{*}{\begin{tabular}[c]{@{}l@{}}Citeseer \\ (3.30\%)\end{tabular}}  & GCond+Ours                 & 72.9\tiny{±0.1} & 71.1\tiny{±0.7} & 71.6\tiny{±1.8}  & 73.4\tiny{±0.4} \\
                                                                               & GCDM+Ours                  & 70.2\tiny{±0.1} & 71.6\tiny{±0.4} & 70.0\tiny{±0.1}  & 71.7\tiny{±0.3} \\
                                                                               & GDEM+Ours                  & 73.2\tiny{±0.2} & 73.3\tiny{±0.4} & 72.3\tiny{±0.5}  & 73.6\tiny{±0.5} \\ \hline
\multirow{3}{*}{\begin{tabular}[c]{@{}l@{}}Pubmed\\ (0.17\%)\end{tabular}}     & GCond+Ours                 & 76.8\tiny{±0.3} & 77.8\tiny{±0.2} & 77.1\tiny{±0.8}  & 75.1\tiny{±1.1} \\
                                                                               & GCDM+Ours                  & 77.3\tiny{±0.4} & 75.0\tiny{±0.1} & 75.6\tiny{±0.5}  & 76.5\tiny{±0.4} \\
                                                                               & GDEM+Ours                  & 78.9\tiny{±0.2} & 78.2\tiny{±0.2} & 77.4\tiny{±0.3}  & 75.7\tiny{±0.8} \\ \hline
\multirow{3}{*}{\begin{tabular}[c]{@{}l@{}}Ogbn-arxiv\\ (0.50\%)\end{tabular}} & GCond+Ours                 & 49.7\tiny{±0.3} & 50.7\tiny{±0.4} & 50.8\tiny{±0.2}  & 49.9\tiny{±0.2} \\
                                                                               & GCDM+Ours                  & 49.3\tiny{±0.4} & 50.7\tiny{±0.3} & 49.2\tiny{±0.2}  & 50.0\tiny{±0.2} \\
                                                                               & GDEM+Ours                  & 52.1\tiny{±0.3} & 52.6\tiny{±0.4} & 51.9\tiny{±0.5}  & 51.5\tiny{±0.4} \\ \hline
\multirow{3}{*}{\begin{tabular}[c]{@{}l@{}}Flickr\\ (0.5\%)\end{tabular}}      & GCond+Ours                 & 48.0\tiny{±0.1} & 47.1\tiny{±0.1} & 46.8\tiny{±0.2}  & 46.7\tiny{±0.1} \\
                                                                               & GCDM+Ours                  & 47.3\tiny{±0.4} & 46.7\tiny{±0.3} & 46.5\tiny{±0.5}  & 46.6\tiny{±0.2} \\
                                                                               & GDEM+Ours                  & 48.5\tiny{±0.1} & 48.7\tiny{±0.2} & 48.1\tiny{±0.3}  & 48.2\tiny{±0.2} \\ \hline
\multirow{3}{*}{\begin{tabular}[c]{@{}l@{}}Reddit\\ (0.1\%)\end{tabular}}      & GCond+Ours                 & 89.3\tiny{±0.7} & 89.4\tiny{±0.5} & 88.9\tiny{±0.3}  & 87.1\tiny{±0.6} \\
                                                                               & GCDM+Ours                  & 89.5\tiny{±0.7} & 88.7\tiny{±0.4} & 89.2\tiny{±0.5}  & 89.0\tiny{±0.4} \\
                                                                               & GDEM+Ours                  & 91.1\tiny{±0.2} & 89.6\tiny{±0.1} & 89.4\tiny{±0.3}  & 90.9\tiny{±0.1} \\ \hline
\multirow{3}{*}{\begin{tabular}[c]{@{}l@{}}Squirrel\\ (1.7\%)\end{tabular}}    & GCond+Ours                 & 24.7\tiny{±1.9} & 25.2\tiny{±1.3} & 23.4\tiny{±2.0}  & 23.3\tiny{±1.4} \\
                                                                               & GCDM+Ours                  & 23.9\tiny{±0.7} & 24.2\tiny{±1.3} & 23.2\tiny{±1.2}  & 23.1\tiny{±1.4} \\
                                                                               & GDEM+Ours                  & 25.5\tiny{±1.9} & 25.2\tiny{±1.5} & 25.5\tiny{±1.2}  & 23.6\tiny{±1.6} \\ \hline
\multirow{3}{*}{\begin{tabular}[c]{@{}l@{}}Gamers\\ (0.5\%)\end{tabular}}      & GCond+Ours                 & 57.8\tiny{±0.1} & 54.2\tiny{±0.3} & 55.1\tiny{±0.4}  & 58.0\tiny{±0.3} \\
                                                                               & GCDM+Ours                  & 58.6\tiny{±0.1} & 54.7\tiny{±0.2} & 54.9\tiny{±0.3}  & 58.2\tiny{±0.2} \\
                                                                               & GDEM+Ours                  & 59.5\tiny{±0.1} & 58.1\tiny{±0.3} & 58.8\tiny{±0.2}  & 59.1\tiny{±0.2} \\ \hline
\end{tabular}}
\end{table}

\subsection{Running Time (Q5)}

The denoising process of RobGC comprises four distinct steps, including: correlation matrix calculation, unreliable edge deletion, locality-based edge addition and parameter search. 
We evaluate the running time for each step and the total denoising and condensation processes in Table \ref{tab_time}. 
Utilizing sparse tensors to store adjacency matrices enables efficient processes for indexing edges and calculating reliability.
Consequently, the running time for the initial three steps is kept under 1 second.
The search process investigates 10 or 20 values for diverse datasets. This results in a total of 100 and 400 threshold combinations being evaluated, respectively.
The table clearly shows that the search times are well-controlled across all datasets.
This efficiency stems primarily from the utilization of sparse matrix multiplication during the label propagation process and the proper configuration of search steps.
The comparison of total running time shows that our method achieves effective denoising results within practical time, making it a viable solution for real-world applications.

\begin{table}[]
\renewcommand{\arraystretch}{1.2}
\setlength{\abovecaptionskip}{0.1cm}
\centering
\caption{The running time (s) for the correlation matrix calculation (CM), unreliable edge deletion (ED), locality-based edge addition (EA), and parameter search (PS) during a single denoising process, as well as the total denoising and condensation time. GCond is deployed to condense the graph. $< 1$ indicates less than 1 second.}
\label{tab_time}
\resizebox{\linewidth}{!}
{
\begin{tabular}{l|cccc}
\hline
\multicolumn{1}{l|}{\begin{tabular}[c]{@{}l@{}}Dataset ($r$)\end{tabular}}       & \multicolumn{1}{l}{\begin{tabular}[c]{@{}l@{}}Cora \\ (5.80\%)\end{tabular}}  & \multicolumn{1}{l}{\begin{tabular}[c]{@{}l@{}}Citeseer \\ (3.30\%)\end{tabular}} & \multicolumn{1}{l}{\begin{tabular}[c]{@{}l@{}}Pubmed\\ (0.17\%)\end{tabular}}  & \multicolumn{1}{l}{\begin{tabular}[c]{@{}l@{}}Ogbn-arxiv\\ (0.50\%)\end{tabular}} \\ \hline
CM+ED+EA     & $< 1$                   & $< 1$                 & $< 1$                & $< 1$                  \\ \hline
PS           & 3.44                     & 3.58                         & 4.59                       & 5.11                           \\ \hline \hline
Denoising    & 107.61                   & 113.25                       & 40.72                      & 112.24                         \\ \hline
Condensation & 498.92                   & 479.31                       & 130.52                     & 12,617.81                      \\ \hline
\end{tabular}}
\end{table}

\begin{table}[t]
\renewcommand{\arraystretch}{1.2}
\centering
\setlength{\abovecaptionskip}{0.1cm}
\caption{{{The ablation study of RobGC. GCond is deployed to condense the graph. Plain indicates the GCond result without denoising for both the training and test graphs. The random noise is applied and the noise level is 100\%. }}}
\label{tab_abl}
\resizebox{\linewidth}{!}
{
\begin{tabular}{l|llll}
\hline
Dataset ($r$) & \begin{tabular}[c]{@{}l@{}}Cora\\ (5.80\%)\end{tabular} & \begin{tabular}[c]{@{}l@{}}Citeseer\\ (3.30\%)\end{tabular} & \begin{tabular}[c]{@{}l@{}}Pubmed\\ (0.17\%)\end{tabular} & \begin{tabular}[c]{@{}l@{}}Ogbn-arxiv\\ (0.50\%)\end{tabular} \\ \hline
Plain         & 64.7±0.8                                                & 66.2±0.1                                                    & 69.8±0.3                                                  & 46.4±0.5                                                      \\
w/o E         & 68.0±0.4                                                & 71.1±0.2                                                    & 74.8±0.4                                                  & 48.2±0.4                                                      \\
w/o propagate & 69.8±0.1                                                & 72.4±0.2                                                    & 76.1±0.4                                                  & 48.9±0.2                                                      \\
w/o addition  & 68.3±0.3                                                & 71.5±0.1                                                    & 75.5±0.1                                                  & 48.1±0.3                                                      \\
w/o deletion  & 68.5±0.3                                                & 71.5±0.3                                                    & 76.2±0.3                                                  & 47.9±0.4                                                      \\ \hline
Ours          & 70.6±0.2                                                & 72.9±0.1                                                    & 76.8±0.3                                                  & 49.7±0.3                                                      \\ \hline
\end{tabular}}
\end{table}

\begin{table*}[t]
\renewcommand{\arraystretch}{1.2}
\centering
\setlength{\abovecaptionskip}{0.1cm}
\caption{{{Comparison between statistics of noisy and optimized graphs. GCond is employed to condense the graph. The random noise is applied and the noise level is 100\%. $\rm{Optimized_d}$ and $\rm{Optimized_a}$ denote the test graph after edge deletion and addition, respectively.}}}
\label{tab_statistic}
\resizebox{\linewidth}{!}
{
\begin{tabular}{l|rrrrr|rrrrr}
\hline
\multirow{3}{*}{Dataset ($r$)}                                                      & \multicolumn{5}{c|}{Cora (5.80\%)}                                                                                                                            & \multicolumn{5}{c}{Citeseer (3.30\%)}                                                                                                                        \\ \cline{2-11} 
                                                                                    & \multicolumn{2}{c|}{Training graph}                        & \multicolumn{3}{c|}{Test graph}                                                                  & \multicolumn{2}{c|}{Training graph}                        & \multicolumn{3}{c}{Test graph}                                                                  \\ \cline{2-11} 
                                                                                    & \multicolumn{1}{c}{Noisy} & \multicolumn{1}{c|}{Optimized} & \multicolumn{1}{c}{Noisy} & \multicolumn{1}{c}{$\rm{Optimized_d}$} & \multicolumn{1}{c|}{$\rm{Optimized_a}$} & \multicolumn{1}{c}{Noisy} & \multicolumn{1}{c|}{Optimized} & \multicolumn{1}{c}{Noisy} & \multicolumn{1}{c}{$\rm{Optimized_d}$} & \multicolumn{1}{c}{$\rm{Optimized_a}$} \\ \hline
\#Nodes                                                                             & 1,028                     & \multicolumn{1}{r|}{1,028}     & 2,208                     & 2,208                            & 2,208                             & 1,827                     & \multicolumn{1}{r|}{1,827}     & 2,827                     & 2,827                            & 2,827                            \\
\#Edges                                                                             & 4,354                     & \multicolumn{1}{r|}{8,993}     & 13,878                    & 11,067                           & 21,636                            & 5,508                     & \multicolumn{1}{r|}{12,756}    & 13,042                    & 12,707                           & 24,442                           \\
Sparsity (\%)                                                                       & 0.30                      & \multicolumn{1}{r|}{0.62}      & 0.28                      & 0.23                             & 0.44                              & 0.17                      & \multicolumn{1}{r|}{0.38}      & 0.16                      & 0.16                             & 0.31                             \\
Homophily                                                                           & 0.52                      & \multicolumn{1}{r|}{0.68}      & 0.50                      & 0.58                             & 0.60                              & 0.42                      & \multicolumn{1}{r|}{0.63}      & 0.45                      & 0.55                             & 0.57                             \\ \hline \hline
\multirow{3}{*}{Dataset ($r$)} & \multicolumn{5}{c|}{Pubmed (0.17\%)}                                                                                                                          & \multicolumn{5}{c}{Ogbn-arxiv (0.50\%)}                                                                                                                      \\ \cline{2-11} 
                                                                                    & \multicolumn{2}{c|}{Training graph}                        & \multicolumn{3}{c|}{Test graph}                                                                  & \multicolumn{2}{c|}{Training graph}                        & \multicolumn{3}{c}{Test graph}                                                                  \\ \cline{2-11} 
                                                                                    & \multicolumn{1}{c}{Noisy} & \multicolumn{1}{c|}{Optimized} & \multicolumn{1}{c}{Noisy} & \multicolumn{1}{c}{$\rm{Optimized_d}$} & \multicolumn{1}{c|}{$\rm{Optimized_a}$} & \multicolumn{1}{c}{Noisy} & \multicolumn{1}{c|}{Optimized} & \multicolumn{1}{c}{Noisy} & \multicolumn{1}{c}{$\rm{Optimized_d}$} & \multicolumn{1}{c}{$\rm{Optimized_a}$} \\ \hline
\#Nodes                                                                             & 18,217                    & \multicolumn{1}{r|}{18,217}    & 19,217                    & 19,217                           & 19,217                            & 90,941                    & \multicolumn{1}{r|}{90,941}    & 139,544                   & 139,544                          & 139,544                          \\
\#Edges                                                                             & 151,014                   & \multicolumn{1}{r|}{200,926}   & 168,544                   & 136,054                          & 219,257                           & 1,404,384                 & \multicolumn{1}{r|}{958,136}   & 3,143,720                 & 1,690,911                        & 2,095,861                        \\
Sparsity (\%)                                                                       & 0.05                      & \multicolumn{1}{r|}{0.06}      & 0.05                      & 0.04                             & 0.06                              & 0.02                      & \multicolumn{1}{r|}{0.01}      & 0.02                      & 0.01                             & 0.01                             \\
Homophily                                                                           & 0.58                      & \multicolumn{1}{r|}{0.77}      & 0.58                      & 0.68                             & 0.74                              & 0.39                      & \multicolumn{1}{r|}{0.63}      & 0.36                      & 0.52                             & 0.54                             \\ \hline
\end{tabular}
}
\end{table*}

\subsection{Ablation Study (Q6)}
To validate the impact of individual components, RobGC is evaluated by disabling specific components, thereby revealing their distinct contributions to the overall performance.
We evaluate RobGC in the following configurations:
(1) ``w/o E'': without calculating correlation matrix $\mathbf{E}$ and leverage the node feature for reliability measurement.
(2) ``w/o propagation'': without propagate the correlation matrix according to condensed graph ${\bf A'}$.
(3) ``w/o addition'': without addition of potential effective edges.
(4) ``w/o deletion'': without deletion of unreliable edges. 
The results of these experimental configurations are summarized in Table \ref{tab_abl}.

The removal of correlation matrix $\mathbf{E}$ leads to a noticeable decline in accuracy, underscoring the importance of global information provided by condensed graphs in structure optimization.
The significance of leveraging the condensed graph is further highlighted by the performance drop in ``w/o propagation''. 
By propagating the correlation matrix across the condensed graph, RobGC can integrate higher-order relationships among condensed nodes, thereby refining the reliability scoring process.
Further analysis of structure optimization strategies highlights the varied significance of edge deletion and addition across different datasets.
Despite their distinct effect, the performance of these two configurations does not match that of the RobGC. This underlines the necessity of integrating both edge deletion and addition strategies to enhance graph quality and improve the performance of RobGC.

\begin{figure}[t]
\setlength{\abovecaptionskip}{0.1cm}
\centering
\includegraphics[width=0.9\linewidth]{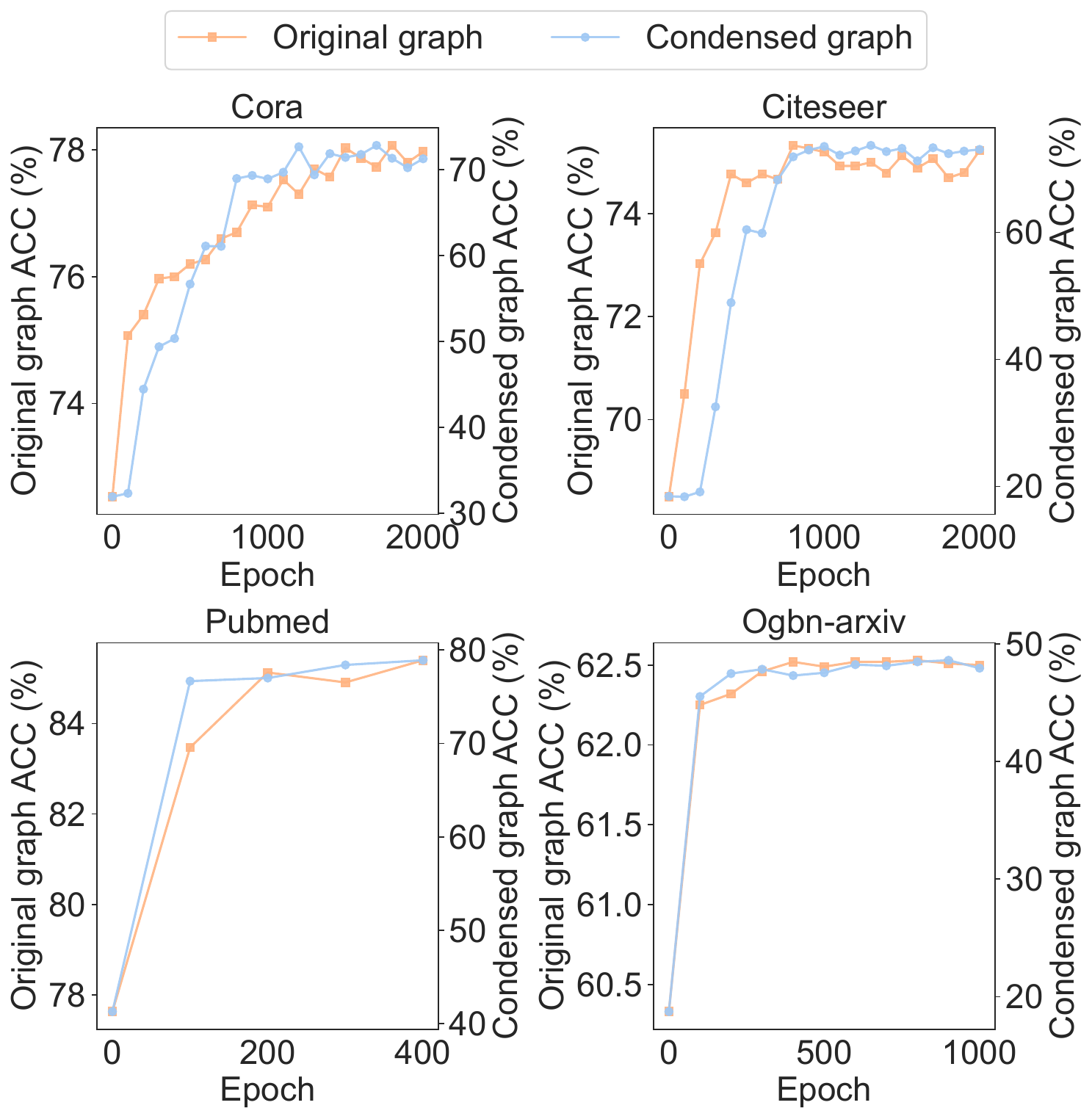}
\caption{{{The accuracy of the model trained on the original and condensed graphs throughout our optimization procedure. The GCond and random noise are applied and the noise level is 100\%. The condensation ratios are 5.80\%, 3.30\%, 0.17\%, and 0.50\% for datasets, respectively.}}}
\label{fig_conver}
\end{figure}

{{
Besides, to verify the convergence of the alternating optimization process, we present the performance of the GCN model trained on both the original and condensed graphs throughout the optimization procedure. The results for all datasets are shown in Fig. \ref{fig_conver}. We observe that the accuracy of the model trained on the original graph gradually improves and converges as the condensation procedure progresses. This demonstrates that the quality of the condensed graph significantly influences the denoising performance of the original graph. Furthermore, as the quality of the original graph improves, the performance of the condensed graph also exhibits gradual enhancement, highlighting the effectiveness of our proposed GC method.
}}

\begin{figure}[t]
\setlength{\abovecaptionskip}{0.1cm}
\centering

\begin{minipage}[t]{0.325\linewidth}
\centering
\includegraphics[width=\linewidth]{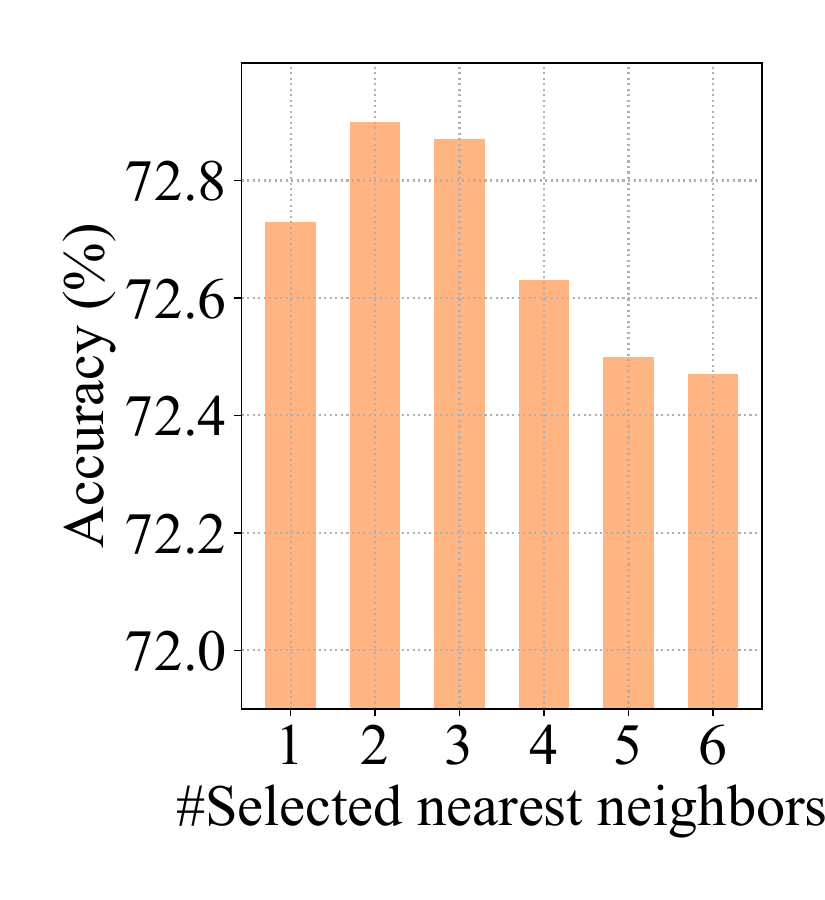}
\end{minipage}
\hfill
\begin{minipage}[t]{0.325\linewidth}
\centering
\includegraphics[width=\linewidth]{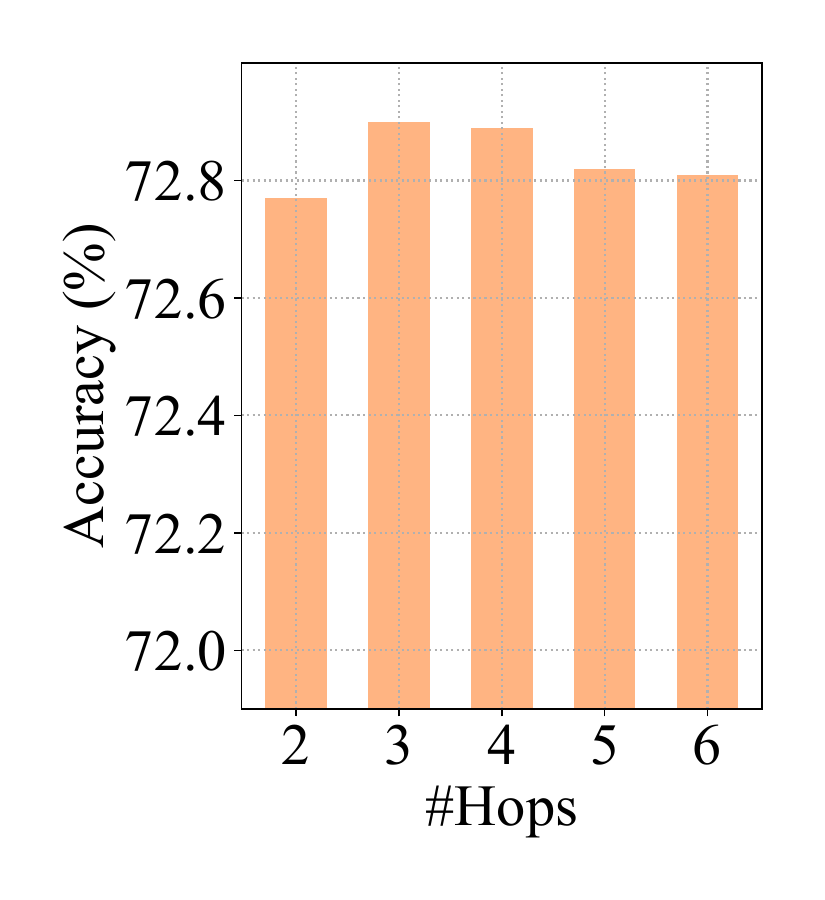}
\end{minipage}
\hfill
\begin{minipage}[t]{0.325\linewidth}
\centering
\includegraphics[width=\linewidth]{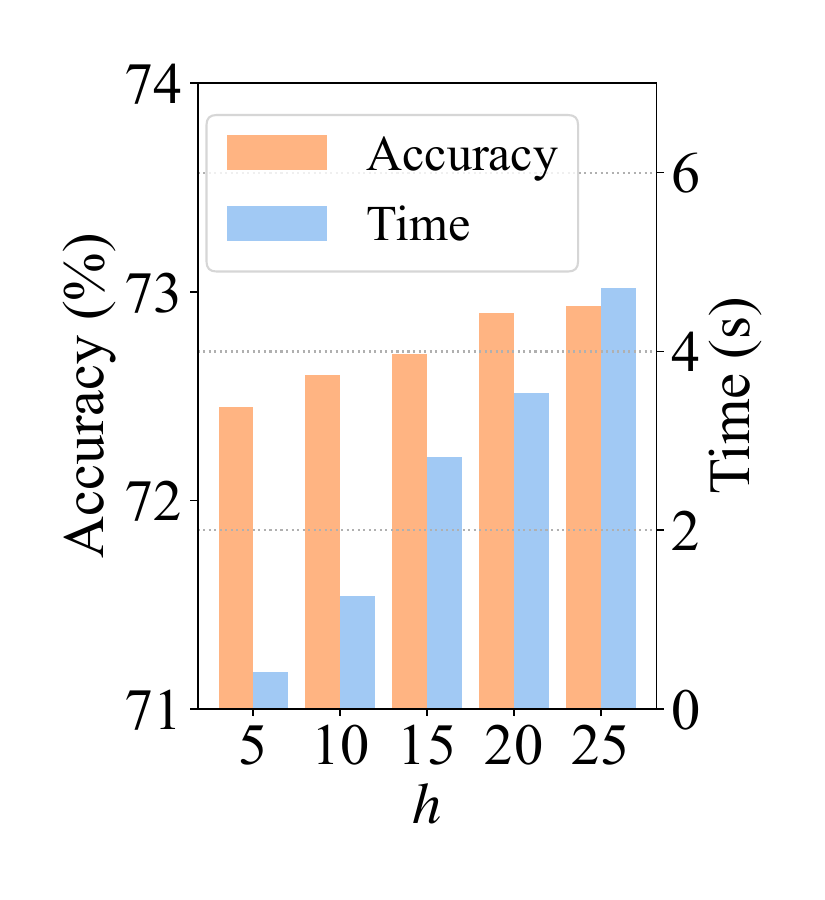}
\end{minipage}

\caption{{{The hyper-parameter sensitivity analysis for the number of selected nearest neighbors, hops, and searching candidate on Citeseer dataset.}}}
\label{fig_hyper}
\end{figure}

\subsection{Analysis on Optimized Graph (Q7)}
To assess the impact of the structure optimization, we compare several properties between optimized graphs and original noisy graphs in Table \ref{tab_statistic}. 
Here, we employ the widely used edge homophily measurement \cite{zhu2020beyond}, which is defined as the fraction of edges in the graph which connects nodes that have the same class label.

The comparison reveals an overall improvement in homophily rates on optimized graphs, with a notable observation that the enhancement in test graphs is not as pronounced as in training graphs. 
The potential rationale is that the thresholds are optimized according to the training graph and directly applied to test data for efficiency. 
{{
Specifically, compared with the noisy test graph, the number of edges decreases after edge deletion and then increases by edge addition. Moreover, homophily consistently increases throughout two phases, demonstrating the effectiveness of both unreliable edge deletion and addition phases.}}
Regarding graph sparsity, we observe an increase in the number of edges within the Cora, Citeseer, and Pubmed datasets. This finding aligns with insights from our ablation study in Table \ref{tab_abl}, indicating that edge addition plays a more crucial role than deletion in these specific graphs.
To obtain the optimal results across different datasets, it's crucial to implement both edge deletion and addition strategies in the structure optimization.

\begin{figure}[t]
\setlength{\abovecaptionskip}{0.1cm}
\centering
\begin{minipage}[t]{0.24\linewidth}
\centering
\includegraphics[width=\linewidth]{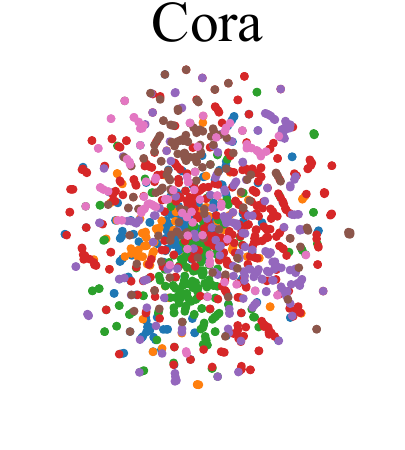}
\end{minipage}
\begin{minipage}[t]{0.24\linewidth}
\centering
\includegraphics[width=\linewidth]{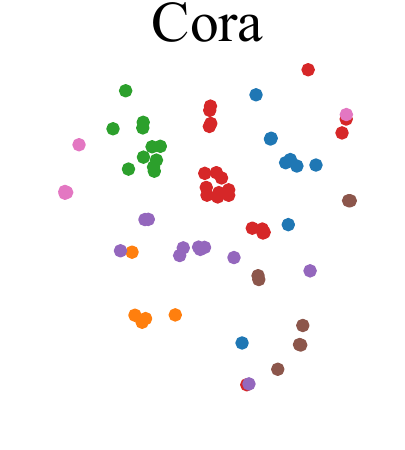}
\end{minipage}
\begin{minipage}[t]{0.24\linewidth}
\centering
\includegraphics[width=\linewidth]{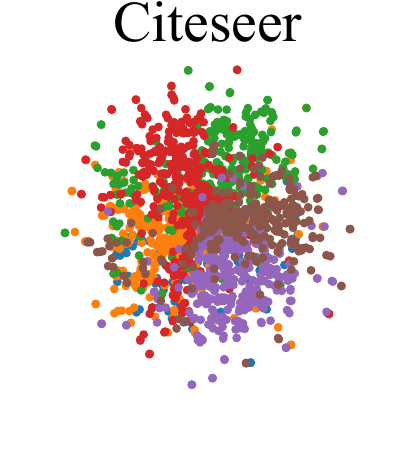}
\end{minipage}
\begin{minipage}[t]{0.24\linewidth}
\centering
\includegraphics[width=\linewidth]{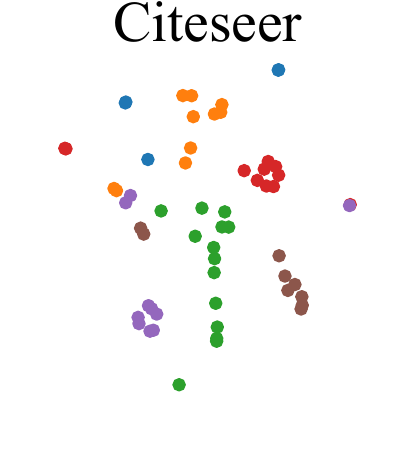}
\end{minipage}
\centering
\begin{minipage}[t]{0.24\linewidth}
\centering
\includegraphics[width=\linewidth]{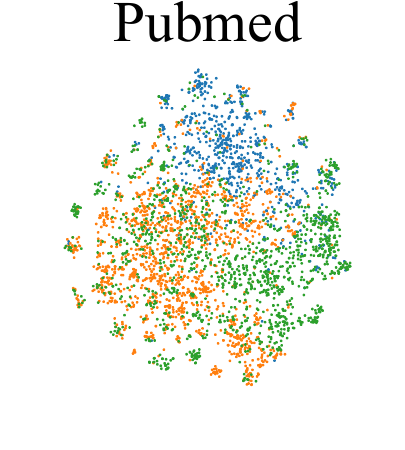}
\end{minipage}
\begin{minipage}[t]{0.24\linewidth}
\centering
\includegraphics[width=\linewidth]{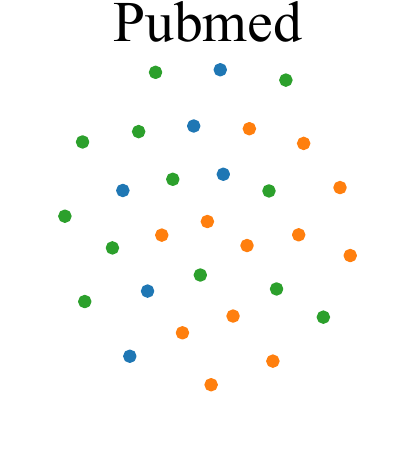}
\end{minipage}
\begin{minipage}[t]{0.24\linewidth}
\centering
\includegraphics[width=\linewidth]{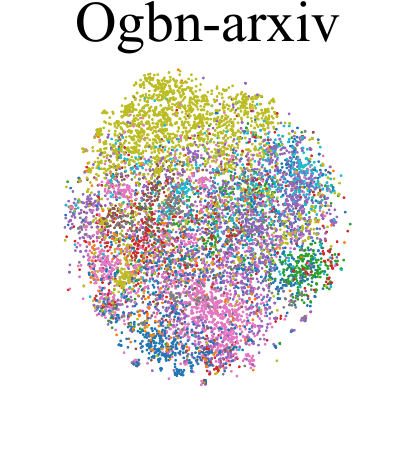}
\end{minipage}
\begin{minipage}[t]{0.24\linewidth}
\centering
\includegraphics[width=\linewidth]{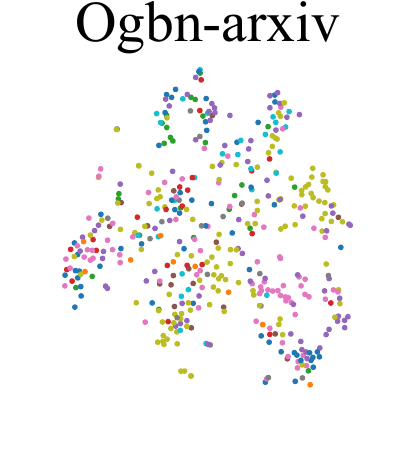}
\end{minipage}
\caption{{{The visualization of the original (left) and condensed (right) graphs, where different colors represent distinct classes.}}}
\label{fig_vis}
\end{figure}

{{In addition, Fig. \ref{fig_vis} presents the t-SNE visualization of node embeddings in the original graph and condensed graph generated by our proposed RobGC. The condensed graph simplifies intra-class distributions while preserving the overall positioning and relationships between classes, ensuring that key inter-class structures are retained. This underscores the representativeness of the condensed node distributions.
}}

\begin{table}[t]
\renewcommand{\arraystretch}{1.2}
\setlength{\abovecaptionskip}{0.1cm}
\centering
\caption{{{The performance of our method combined with robust GNNs. The random noise is applied and the noise level is 100\%. The condensation ratios are 5.80\%, 3.30\%, 0.17\%, and 0.50\% for datasets, respectively.}}}
\label{tabrobustgnn}
\resizebox{\linewidth}{!}
{
\begin{tabular}{l|cccc}
\hline
                        & Cora     & Citeseer & Pubmed   & Ogbn-arxiv \\ \hline
GCond+GCN               & 64.7±0.8 & 66.2±0.1 & 69.8±0.3 & 46.4±0.5   \\ \hline
GCond+${\rm{STABLE_M}}$ & 65.5±0.7 & 67.6±0.2 & 70.2±0.5 & 47.5±0.6   \\
Ours+${\rm{STABLE_M}}$  & 70.9±0.3 & 73.2±0.1 & 77.0±0.3 & 50.1±0.3   \\ \hline
GCond+Soft Median       & 65.8±0.5 & 67.9±0.4 & 70.8±0.3 & 47.9±0.5   \\
Ours+Soft Median        & 71.2±0.4 & 73.9±0.3 & 77.3±0.2 & 50.6±0.3   \\ \hline
\end{tabular}}
\end{table}

\subsection{Hyper-parameter Sensitivity Analysis (Q8)}
In this section, we evaluate the number of selected nearest neighbors and hops for neighbor candidates in the locality-based edge addition, as well as the number of searching candidates $h$ for optimization.
In Fig. \ref{fig_hyper}, we present the performances on Citeseer and similar trends were noted across other datasets.
For the number of selected nearest neighbors, we observe that choosing 2 to 3 nearest neighbors yields the best performance. Higher values may include an excessive number of neighbors, resulting in a denser adjacency matrix that could potentially introduce more noisy edges.
On the other hand, the performance shows relative insensitivity to the number of hops for neighbor candidates.
While a higher number of hops can encompass more potentially effective neighbors, it also significantly increases the number of candidates and computational complexity.
{{
For the number of searching candidates, search time increase with larger values of $h$, and accuracy levels off beyond a certain point.
Therefore, we select the number of hop as 3 and $h$ as 20 in the experiments to maintain a balance between performance and computation demands.}}

\subsection{{{Performance for Robust GNNs (Q9)}}}
{{To further evaluate the generalizability of our method, we evaluate two robust GNNs: ${\rm{STABLE_M}}$~\cite{li2022reliable} and Soft Median~\cite{geisler2021robustness}, on the denoised condensed graphs. Table \ref{tabrobustgnn} compares data-centric methods for generating condensed graphs (GCond and Ours) and various GNN models (GCN, ${\rm{STABLE_M}}$, and Soft Median).  
The results show that robust GNNs outperform the vanilla GCN model across different datasets. However, their performance is still impacted by the noisy condensed graph. When combined with our RobGC, their performance improves significantly, underscoring the effectiveness of our denoising strategy.}}

\section{Related Work}

\subsection{Graph Condensation}

{{
Existing graph condensation \cite{gao2024graph, ijcai2024p891} research can be broadly categorized into 4 classes based on the primary criteria: effective GC, generalized GC, fair GC, and efficient GC. 
(1) Effective GC prioritizes improving the accuracy of GNNs trained on condensed graphs. 
These approaches often employ advanced optimization strategies to enhance the optimization results, such as trajectory matching \cite{zheng_structure_free_2023}, kernel ridge regression \cite{jin2022condensing}, and graph neural tangent kernels \cite{zheng_structure_free_2023}. Moreover, recent studies explore the advanced techniques to explore the inherent information in the original graph, including self-supervised learning \cite{10.1145/3637528.3671682}, difficult sample mining \cite{zhang2024navigating}, and self-expressive ability \cite{liu2024graphself}.
(2) Generalized GC \cite{yang_does_2023} aims to improve the overall performance of different GNN architectures trained on the condensed graph. The primary challenge of generalization is to minimize the loss of information in the structure during condensation. 
(3) Fair GC \cite{feng_fair_2023} focuses on the disparity in fairness between models trained on condensed graphs and those trained on training graphs, aiming to prevent bias amplification in the condensed graph and ensure equitable model performance. 
(4) Efficient GC \cite{zhao2023dataset} concentrates on the speed of graph generation and the method’s feasibility for time-sensitive scenarios. These methods are developed to accelerate the condensation process across all components in GC, including: graph encoding \cite{gao2024grapho}, optimization \cite{gao2024rethinking}, and graph generation \cite{xiao2024simple}.}}

{{Besides these methods, another related GC study is MCond~\cite{gao_graph_2023}, which addresses the high latency issue in the same inductive inference setting as ours. It introduces a mapping matrix that connects the inductive graph to the condensed graph for efficient message-passing. Building on this insight, RobGC can be further refined to enable inductive graphs to perform efficient message-passing on noise-free condensed graphs, with significant potential to enhance GC performance in both resource-constrained, and noisy scenarios.}}

\subsection{Graph Structure Learning}
To enhance the robustness of GNN models, graph structure learning has been introduced to jointly optimize the graph structure while learning corresponding node representations, which can be categorized into three classes: metric learning, probabilistic modeling, and direct optimization. 

Specifically, metric learning approaches model edge weights as the distance between each pair of node representations and introduces numerous metric functions like Gaussian kernel \cite{li2018adaptive}, inner product \cite{yu2021graph}, and cosine similarity \cite{zhang2020gnnguard}.
Each metric offers a distinct approach to defining the similarity between nodes, thereby guiding the formation of edges and influencing the graph structure optimization process.
Probabilistic modeling approaches \cite{franceschi2019learning,luo2021learning} assume that the graph is generated by a specific distribution. Consequently, these approaches use a trainable model to define the edge probability distribution and apply re-parameterization techniques to learn the discrete graph structure.
Direct optimization approaches \cite{yang2019topology,wan2021graph} consider the optimized graph adjacency matrix as learnable parameters and optimize the adjacency matrix toward achieving a feasible graph structure. Therefore, they employ regularization techniques informed by prior knowledge of graphs, emphasizing aspects like feature smoothness \cite{wu2018quest}, sparsity \cite{chen2020iterative}, and the low-rank nature \cite{jin2020graph} of the graph structure.

\section{Conclusion}
In this paper, we present robust graph condensation (RobGC), a general approach for GC to extend the robustness and applicability of condensed graphs in noisy scenarios.
RobGC incorporates a denoising process to tackle noise present in the training graph and adopts an alternating optimization strategy for condensation and denoising processes, contributing to the mutual promotion of the condensed graph and the training graph.
Additionally, RobGC facilitates the inductive test-time graph denoising by leveraging the noise-free condensed graph, preventing the inductive nodes from noise during inference.
{{Our empirical analysis demonstrates that adversarial attacks pose a substantial threat to the quality of graph structures. Future research could further investigate the advanced relay model in GC, incorporating robust aggregation functions~\cite{geisler2020reliable}  to filter out noise and counteract the effects of adversarial attacks on both features and labels.}}

\bibliographystyle{IEEEtran}
\bibliography{IEEEabrv,ref2}

\begin{thebibliography}{10}
\providecommand{\url}[1]{#1}
\csname url@samestyle\endcsname
\providecommand{\newblock}{\relax}
\providecommand{\bibinfo}[2]{#2}
\providecommand{\BIBentrySTDinterwordspacing}{\spaceskip=0pt\relax}
\providecommand{\BIBentryALTinterwordstretchfactor}{4}
\providecommand{\BIBentryALTinterwordspacing}{\spaceskip=\fontdimen2\font plus
\BIBentryALTinterwordstretchfactor\fontdimen3\font minus \fontdimen4\font\relax}
\providecommand{\BIBforeignlanguage}[2]{{%
\expandafter\ifx\csname l@#1\endcsname\relax
\typeout{** WARNING: IEEEtran.bst: No hyphenation pattern has been}%
\typeout{** loaded for the language `#1'. Using the pattern for}%
\typeout{** the default language instead.}%
\else
\language=\csname l@#1\endcsname
\fi
#2}}
\providecommand{\BIBdecl}{\relax}
\BIBdecl

\bibitem{zheng2016keyword}
B.~Zheng, K.~Zheng, X.~Xiao, H.~Su, H.~Yin, X.~Zhou, and G.~Li, ``Keyword-aware continuous knn query on road networks,'' in \emph{IEEE 32nd International Conference on Data Engineering}, 2016, pp. 871--882.

\bibitem{gao2023semantic}
X.~Gao, W.~Zhang, T.~Chen, J.~Yu, H.~Q.~V. Nguyen, and H.~Yin, ``Semantic-aware node synthesis for imbalanced heterogeneous information networks,'' in \emph{Proceedings of the 32nd ACM International Conference on Information and Knowledge Management}, 2023, pp. 545--555.

\bibitem{gao2023accelerating}
X.~Gao, W.~Zhang, J.~Yu, Y.~Shao, Q.~V.~H. Nguyen, B.~Cui, and H.~Yin, ``Accelerating scalable graph neural network inference with node-adaptive propagation,'' in \emph{IEEE 40th International Conference on Data Engineering}, 2024, pp. 3042--3055.

\bibitem{guo2022graph}
Z.~Guo, K.~Guo, B.~Nan, Y.~Tian, R.~G. Iyer, Y.~Ma, O.~Wiest, X.~Zhang, W.~Wang, C.~Zhang \emph{et~al.}, ``Graph-based molecular representation learning,'' \emph{arXiv preprint arXiv:2207.04869}, 2022.

\bibitem{li2018influence}
Y.~Li, J.~Fan, Y.~Wang, and K.-L. Tan, ``Influence maximization on social graphs: A survey,'' \emph{IEEE Transactions on Knowledge and Data Engineering}, vol.~30, no.~10, pp. 1852--1872, 2018.

\bibitem{jiang2024challenging}
W.~Jiang, X.~Gao, G.~Xu, T.~Chen, and H.~Yin, ``Challenging low homophily in social recommendation,'' in \emph{Proceedings of the ACM Web Conference}, 2024, pp. 3476--3484.

\bibitem{DBLP:conf/iclr/ZengZSKP20}
H.~Zeng, H.~Zhou, A.~Srivastava, R.~Kannan, and V.~Prasanna, ``{GraphSAINT}: Graph sampling based inductive learning method,'' in \emph{International Conference on Learning Representations}, 2020.

\bibitem{zhang2022pasca}
W.~Zhang, Y.~Shen, Z.~Lin, Y.~Li, X.~Li, W.~Ouyang, Y.~Tao, Z.~Yang, and B.~Cui, ``{PaSca}: A graph neural architecture search system under the scalable paradigm,'' in \emph{Proceedings of the ACM Web Conference}, 2022, pp. 1817--1828.

\bibitem{rebuffi2017icarl}
S.-A. Rebuffi, A.~Kolesnikov, G.~Sperl, and C.~H. Lampert, ``{iCaRL}: Incremental classifier and representation learning,'' in \emph{Proceedings of the IEEE Conference on Computer Vision and Pattern Recognition}, 2017, pp. 5533--5542.

\bibitem{pan_fedgkd_2023}
Q.~Pan, R.~Wu, T.~Liu, T.~Zhang, Y.~Zhu, and W.~Wang, ``{FedGKD}: Unleashing the power of collaboration in federated graph neural networks,'' \emph{arXiv preprint arXiv:2309.09517}, 2023.

\bibitem{jin2022graph}
W.~Jin, L.~Zhao, S.~Zhang, Y.~Liu, J.~Tang, and N.~Shah, ``Graph condensation for graph neural networks,'' in \emph{International Conference on Learning Representations}, 2022.

\bibitem{gao2024graph}
X.~Gao, J.~Yu, T.~Chen, G.~Ye, W.~Zhang, and H.~Yin, ``Graph condensation: A survey,'' \emph{IEEE Transactions on Knowledge and Data Engineering}, 2025.

\bibitem{van2022inductive}
R.~{Van Belle}, C.~{Van Damme}, H.~Tytgat, and J.~{De Weerdt}, ``Inductive graph representation learning for fraud detection,'' \emph{Expert Systems with Applications}, vol. 193, p. 116463, 2022.

\bibitem{hung2017computing}
N.~Q.~V. Hung, H.~H. Viet, N.~T. Tam, M.~Weidlich, H.~Yin, and X.~Zhou, ``Computing crowd consensus with partial agreement,'' \emph{IEEE Transactions on Knowledge and Data Engineering}, vol.~30, no.~1, pp. 1--14, 2018.

\bibitem{nguyen2017argument}
Q.~V.~H. Nguyen, C.~T. Duong, T.~T. Nguyen, M.~Weidlich, K.~Aberer, H.~Yin, and X.~Zhou, ``Argument discovery via crowdsourcing,'' \emph{The VLDB Journal}, vol.~26, pp. 511--535, 2017.

\bibitem{wang2018streaming}
W.~Wang, H.~Yin, Z.~Huang, Q.~Wang, X.~Du, and Q.~V.~H. Nguyen, ``Streaming ranking based recommender systems,'' in \emph{The 41st International ACM SIGIR Conference on Research \& Development in Information Retrieval}, 2018, pp. 525--534.

\bibitem{hamilton2017inductive}
W.~Hamilton, Z.~Ying, and J.~Leskovec, ``Inductive representation learning on large graphs,'' in \emph{Advances in Neural Information Processing Systems}, vol.~30, 2017.

\bibitem{DBLP:conf/iclr/KipfW17}
T.~N. Kipf and M.~Welling, ``Semi-supervised classification with graph convolutional networks,'' in \emph{International Conference on Learning Representations}, 2017.

\bibitem{randall2014use}
S.~M. Randall, J.~H. Boyd, A.~M. Ferrante, J.~K. Bauer, and J.~B. Semmens, ``Use of graph theory measures to identify errors in record linkage,'' \emph{Computer Methods and Programs in Biomedicine}, vol. 115, no.~2, pp. 55--63, 2014.

\bibitem{tu2023deep}
H.~Tu, S.~Yu, V.~Saikrishna, F.~Xia, and K.~Verspoor, ``Deep outdated fact detection in knowledge graphs,'' in \emph{IEEE International Conference on Data Mining Workshops}, 2023, pp. 1443--1452.

\bibitem{gao2021training}
Z.~Gao, S.~Bhattacharya, L.~Zhang, R.~S. Blum, A.~Ribeiro, and B.~M. Sadler, ``Training robust graph neural networks with topology adaptive edge dropping,'' \emph{arXiv preprint arXiv:2106.02892}, 2021.

\bibitem{geisler2021robustness}
S.~Geisler, T.~Schmidt, H.~{\c{S}}irin, D.~Z{\"u}gner, A.~Bojchevski, and S.~G{\"u}nnemann, ``Robustness of graph neural networks at scale,'' in \emph{Advances in Neural Information Processing Systems}, vol.~34, 2021, pp. 7637--7649.

\bibitem{zhu2021deep}
Y.~Zhu, W.~Xu, J.~Zhang, Q.~Liu, S.~Wu, and L.~Wang, ``Deep graph structure learning for robust representations: A survey,'' \emph{arXiv preprint arXiv:2103.03036}, 2021.

\bibitem{geisler2020reliable}
S.~Geisler, D.~Z{\"u}gner, and S.~G{\"u}nnemann, ``Reliable graph neural networks via robust aggregation,'' in \emph{Advances in Neural Information Processing Systems}, vol.~33, 2020, pp. 13\,272--13\,284.

\bibitem{li2022reliable}
K.~Li, Y.~Liu, X.~Ao, J.~Chi, J.~Feng, H.~Yang, and Q.~He, ``Reliable representations make a stronger defender: Unsupervised structure refinement for robust gnn,'' in \emph{Proceedings of the 28th ACM SIGKDD Conference on Knowledge Discovery and Data Mining}, 2022, pp. 925--935.

\bibitem{lei_comprehensive_2024}
S.~Lei and D.~Tao, ``A comprehensive survey of dataset distillation,'' \emph{IEEE Transactions on Pattern Analysis \& Machine Intelligence}, vol.~46, no.~01, pp. 17--32, 2024.

\bibitem{zhao2023dataset}
B.~Zhao and H.~Bilen, ``Dataset condensation with distribution matching,'' in \emph{Proceedings of the IEEE/CVF Winter Conference on Applications of Computer Vision}, 2023, pp. 6514--6523.

\bibitem{liu_cat_2023}
Y.~Liu, R.~Qiu, and Z.~Huang, ``{CaT}: Balanced continual graph learning with graph condensation,'' in \emph{IEEE International Conference on Data Mining}, 2023, pp. 1157--1162.

\bibitem{liu_graph_2023}
Y.~Liu, D.~Bo, and C.~Shi, ``Graph distillation with eigenbasis matching,'' in \emph{Proceedings of the 41st International Conference on Machine Learning}, vol. 235, 2024, pp. 30\,702--30\,717.

\bibitem{wang2023fast}
L.~Wang, W.~Fan, J.~Li, Y.~Ma, and Q.~Li, ``Fast graph condensation with structure-based neural tangent kernel,'' in \emph{Proceedings of the ACM Web Conference}, 2024, pp. 4439--4448.

\bibitem{gao2024grapho}
X.~Gao, T.~Chen, W.~Zhang, Y.~Li, X.~Sun, and H.~Yin, ``Graph condensation for open-world graph learning,'' in \emph{Proceedings of the 30th ACM SIGKDD Conference on Knowledge Discovery and Data Mining}, 2024, pp. 851--862.

\bibitem{zheng_structure_free_2023}
X.~Zheng, M.~Zhang, C.~Chen, Q.~V.~H. Nguyen, X.~Zhu, and S.~Pan, ``Structure-free graph condensation: From large-scale graphs to condensed graph-free data,'' in \emph{Advances in Neural Information Processing Systems}, vol.~36, 2023, pp. 6026--6047.

\bibitem{zhang2024navigating}
Y.~Zhang, T.~Zhang, K.~Wang, Z.~Guo, Y.~Liang, X.~Bresson, W.~Jin, and Y.~You, ``Navigating complexity: Toward lossless graph condensation via expanding window matching,'' in \emph{Proceedings of the 41st International Conference on Machine Learning}, vol. 235, 2024, pp. 60\,379--60\,395.

\bibitem{liu2022towards}
Y.~Liu, Y.~Zheng, D.~Zhang, H.~Chen, H.~Peng, and S.~Pan, ``Towards unsupervised deep graph structure learning,'' in \emph{Proceedings of the ACM Web Conference}, 2022, pp. 1392--1403.

\bibitem{wang2021combining}
H.~Wang and J.~Leskovec, ``Combining graph convolutional neural networks and label propagation,'' \emph{ACM Transactions on Information Systems}, vol.~40, no.~4, pp. 1--27, 2021.

\bibitem{lu2022ensemble}
Y.~Lu, Y.~Bo, and W.~He, ``An ensemble model for combating label noise,'' in \emph{Proceedings of the Fifteenth ACM International Conference on Web Search and Data Mining}, 2022, pp. 608--617.

\bibitem{tan2021co}
C.~Tan, J.~Xia, L.~Wu, and S.~Z. Li, ``Co-learning: Learning from noisy labels with self-supervision,'' in \emph{Proceedings of the 29th ACM International Conference on Multimedia}, 2021, pp. 1405--1413.

\bibitem{yuan2024hide}
B.~Yuan, Y.~Chen, Y.~Zhang, and W.~Jiang, ``Hide and seek in noise labels: Noise-robust collaborative active learning with {LLM}s-powered assistance,'' in \emph{Proceedings of the 62nd Annual Meeting of the Association for Computational Linguistics}, 2024, pp. 10\,977--11\,011.

\bibitem{page1999pagerank}
L.~Page, S.~Brin, R.~Motwani, and T.~Winograd, ``The pagerank citation ranking: Bringing order to the web.'' Stanford InfoLab, Tech. Rep., 1999.

\bibitem{jin2020graph}
W.~Jin, Y.~Ma, X.~Liu, X.~Tang, S.~Wang, and J.~Tang, ``Graph structure learning for robust graph neural networks,'' in \emph{Proceedings of the 26th ACM SIGKDD International Conference on Knowledge Discovery \& Data Mining}, 2020, pp. 66--74.

\bibitem{jin2021node}
W.~Jin, T.~Derr, Y.~Wang, Y.~Ma, Z.~Liu, and J.~Tang, ``Node similarity preserving graph convolutional networks,'' in \emph{Proceedings of the 14th ACM International Conference on Web Search and Data Mining}, 2021, pp. 148--156.

\bibitem{wu2019simplifying}
F.~Wu, A.~H.~S. Jr., T.~Zhang, C.~Fifty, T.~Yu, and K.~Q. Weinberger, ``Simplifying graph convolutional networks,'' in \emph{Proceedings of the 36th International Conference on Machine Learning}, vol.~97, 2019, pp. 6861--6871.

\bibitem{klicpera2018predict}
J.~Klicpera, A.~Bojchevski, and S.~G{\"u}nnemann, ``Predict then propagate: Graph neural networks meet personalized pagerank,'' \emph{arXiv preprint arXiv:1810.05997}, 2018.

\bibitem{zhu2020beyond}
J.~Zhu, Y.~Yan, L.~Zhao, M.~Heimann, L.~Akoglu, and D.~Koutra, ``Beyond homophily in graph neural networks: Current limitations and effective designs,'' in \emph{Advances in Neural Information Processing Systems}, vol.~33, 2020, pp. 7793--7804.

\bibitem{ijcai2024p891}
M.~Hashemi, S.~Gong, J.~Ni, W.~Fan, B.~A. Prakash, and W.~Jin, ``A comprehensive survey on graph reduction: Sparsification, coarsening, and condensation,'' in \emph{Proceedings of the Thirty-Third International Joint Conference on Artificial Intelligence}, 2024, pp. 8058--8066, survey Track.

\bibitem{jin2022condensing}
W.~Jin, X.~Tang, H.~Jiang, Z.~Li, D.~Zhang, J.~Tang, and B.~Yin, ``Condensing graphs via one-step gradient matching,'' in \emph{Proceedings of the 28th ACM SIGKDD Conference on Knowledge Discovery and Data Mining}, 2022, pp. 720--730.

\bibitem{10.1145/3637528.3671682}
Y.~Wang, X.~Yan, S.~Jin, H.~Huang, Q.~Xu, Q.~Zhang, B.~Du, and J.~Jiang, ``Self-supervised learning for graph dataset condensation,'' in \emph{Proceedings of the 30th ACM SIGKDD Conference on Knowledge Discovery and Data Mining}, 2024, pp. 3289--3298.

\bibitem{liu2024graphself}
Z.~Liu, C.~Zeng, and G.~Zheng, ``Graph data condensation via self-expressive graph structure reconstruction,'' in \emph{Proceedings of the 30th ACM SIGKDD Conference on Knowledge Discovery and Data Mining}, 2024, pp. 1992--2002.

\bibitem{yang_does_2023}
B.~Yang, K.~Wang, Q.~Sun, C.~Ji, X.~Fu, H.~Tang, Y.~You, and J.~Li, ``Does graph distillation see like vision dataset counterpart?'' in \emph{Advances in Neural Information Processing Systems}, vol.~36, 2023, pp. 53\,201--53\,226.

\bibitem{feng_fair_2023}
Q.~Feng, Z.~Jiang, R.~Li, Y.~Wang, N.~Zou, J.~Bian, and X.~Hu, ``Fair graph distillation,'' in \emph{Advances in Neural Information Processing Systems}, vol.~36, 2023, pp. 80\,644--80\,660.

\bibitem{gao2024rethinking}
X.~Gao, T.~Chen, W.~Zhang, J.~Yu, G.~Ye, Q.~V.~H. Nguyen, and H.~Yin, ``Rethinking and accelerating graph condensation: A training-free approach with class partition,'' in \emph{Proceedings of the ACM Web Conference}, 2025.

\bibitem{xiao2024simple}
Z.~Xiao, Y.~Wang, S.~Liu, H.~Wang, M.~Song, and T.~Zheng, ``Simple graph condensation,'' in \emph{Machine Learning and Knowledge Discovery in Databases. Research Track: European Conference, ECML PKDD}, 2024, pp. 53--71.

\bibitem{gao_graph_2023}
X.~Gao, T.~Chen, Y.~Zang, W.~Zhang, Q.~V. Hung~Nguyen, K.~Zheng, and H.~Yin, ``{Graph Condensation for Inductive Node Representation Learning},'' in \emph{IEEE 40th International Conference on Data Engineering}, 2024, pp. 3056--3069.

\bibitem{li2018adaptive}
R.~Li, S.~Wang, F.~Zhu, and J.~Huang, ``Adaptive graph convolutional neural networks,'' in \emph{Proceedings of the AAAI Conference on Artificial Intelligence}, vol.~32, no.~1, 2018.

\bibitem{yu2021graph}
D.~Yu, R.~Zhang, Z.~Jiang, Y.~Wu, and Y.~Yang, ``Graph-revised convolutional network,'' in \emph{Machine Learning and Knowledge Discovery in Databases: European Conference, ECML PKDD}, 2020, pp. 378--393.

\bibitem{zhang2020gnnguard}
X.~Zhang and M.~Zitnik, ``{GNNGuard}: Defending graph neural networks against adversarial attacks,'' in \emph{Advances in Neural Information Processing Systems}, H.~Larochelle, M.~Ranzato, R.~Hadsell, M.~Balcan, and H.~Lin, Eds., vol.~33.\hskip 1em plus 0.5em minus 0.4em\relax Curran Associates, Inc., 2020, pp. 9263--9275.

\bibitem{franceschi2019learning}
L.~Franceschi, M.~Niepert, M.~Pontil, and X.~He, ``Learning discrete structures for graph neural networks,'' in \emph{Proceedings of the 36th International Conference on Machine Learning}, vol.~97, 2019, pp. 1972--1982.

\bibitem{luo2021learning}
D.~Luo, W.~Cheng, W.~Yu, B.~Zong, J.~Ni, H.~Chen, and X.~Zhang, ``Learning to drop: Robust graph neural network via topological denoising,'' in \emph{Proceedings of the 14th ACM International Conference on Web Search and Data Mining}, 2021, pp. 779--787.

\bibitem{yang2019topology}
L.~Yang, Z.~Kang, X.~Cao, D.~Jin, B.~Yang, and Y.~Guo, ``Topology optimization based graph convolutional network,'' in \emph{Proceedings of the Twenty-Eighth International Joint Conference on Artificial Intelligence}, 2019, pp. 4054--4061.

\bibitem{wan2021graph}
G.~Wan and H.~Kokel, ``Graph sparsification via meta-learning,'' in \emph{DLG@ AAAI}, 2021.

\bibitem{wu2018quest}
X.~Wu, L.~Zhao, and L.~Akoglu, ``A quest for structure: Jointly learning the graph structure and semi-supervised classification,'' in \emph{Proceedings of the 27th ACM International Conference on Information and Knowledge Management}, 2018, pp. 87--96.

\bibitem{chen2020iterative}
Y.~Chen, L.~Wu, and M.~Zaki, ``Iterative deep graph learning for graph neural networks: Better and robust node embeddings,'' in \emph{Advances in Neural Information Processing Systems}, vol.~33, 2020, pp. 19\,314--19\,326.

\end{thebibliography}

\end{document}